\pdfoutput=1

\documentclass[11pt]{article}

\usepackage{ACL2023}

\usepackage{times}
\usepackage{latexsym}

\usepackage[T1]{fontenc}

\usepackage[utf8]{inputenc}

\usepackage{microtype}

\usepackage{inconsolata}

\usepackage{hyperref}       
\usepackage{url}            
\usepackage{booktabs}       
\usepackage{amsfonts}       
\usepackage{nicefrac}       
\usepackage{microtype}      
\usepackage{makecell}
\usepackage{graphicx}
\usepackage{subfig}
\usepackage{float}
\usepackage{multirow}
\usepackage{bm}
\usepackage{amsopn}
\usepackage{amsmath}
\usepackage{xcolor}
\usepackage{color}
\usepackage{wrapfig}
\usepackage{multibib}
\usepackage{booktabs}
\usepackage{verbatim}
\usepackage{colortbl}
\usepackage{makecell}
\usepackage{amssymb}
\usepackage{bbm}
\usepackage{algorithm}
\usepackage{algorithmic}
\usepackage{pifont}

\usepackage{hyperref}
\makeatletter
\def\UrlAlphabet{%
      \do\a\do\b\do\c\do\d\do\e\do\f\do\g\do\h\do\i\do\j%
      \do\k\do\l\do\m\do\n\do\o\do\p\do\q\do\r\do\s\do\t%
      \do\u\do\v\do\w\do\x\do\y\do\z\do\A\do\B\do\C\do\D%
      \do\E\do\F\do\G\do\H\do\I\do\J\do\K\do\L\do\M\do\N%
      \do\O\do\P\do\Q\do\R\do\S\do\T\do\U\do\V\do\W\do\X%
      \do\Y\do\Z}
\def\UrlDigits{\do\1\do\2\do\3\do\4\do\5\do\6\do\7\do\8\do\9\do\0}
\g@addto@macro{\UrlBreaks}{\UrlOrds}
\g@addto@macro{\UrlBreaks}{\UrlAlphabet}
\g@addto@macro{\UrlBreaks}{\UrlDigits}
\makeatother

\title{
How Do In-Context Examples Affect Compositional Generalization?
}

\author{
  Shengnan An\thanks{\, Work done during an internship at Microsoft Research.}\hspace{0.4mm} $^{\dagger}$, Zeqi Lin$^{\ddagger}$, Qiang Fu$^{\ddagger}$, Bei Chen$^{\ddagger}$, \\
  \textbf{Nanning Zheng$^{\dagger}$, Jian-Guang LOU$^{\ddagger}$, Dongmei Zhang$^{\ddagger}$}\\
  $^{\dagger}$ Institute of Artificial Intelligence and Robotics, Xi’an Jiaotong University\\
  $^{\ddagger}$ Microsoft Corporation\\
  \texttt{\{an1006634493@stu, nnzheng@mail\}.xjtu.edu.cn} \\
  \texttt{\{Zeqi.Lin, qifu, beichen, jlou, dongmeiz\}@microsoft.com}
}

\begin{document}
\maketitle
\begin{abstract}
Compositional generalization\textemdash understanding unseen combinations of seen primitives\textemdash is an essential reasoning capability in human intelligence.
The AI community mainly studies this capability by fine-tuning neural networks on lots of training samples, while it is still unclear whether and how in-context learning\textemdash the prevailing few-shot paradigm based on large language models\textemdash exhibits compositional generalization.
In this paper, we present \textsc{CoFe}, a test suite to investigate in-context compositional generalization.
We find that the compositional generalization performance can be easily affected by the selection of in-context examples, thus raising the research question what the key factors are to make good in-context examples for compositional generalization.
We study three potential factors: similarity, diversity and complexity. Our systematic experiments indicate that in-context examples should be structurally similar to the test case, diverse from each other, and individually simple.
Furthermore, two strong limitations are observed: in-context compositional generalization on fictional words is much weaker than that on commonly used ones; it is still critical that the in-context examples should cover required linguistic structures, even though the backbone model has been pre-trained on large corpus.
We hope our analysis would facilitate the understanding and utilization of in-context learning paradigm.

\end{abstract}

\section{Introduction}

\begin{figure}[t]
    \centering
    \includegraphics[width=.47\textwidth]{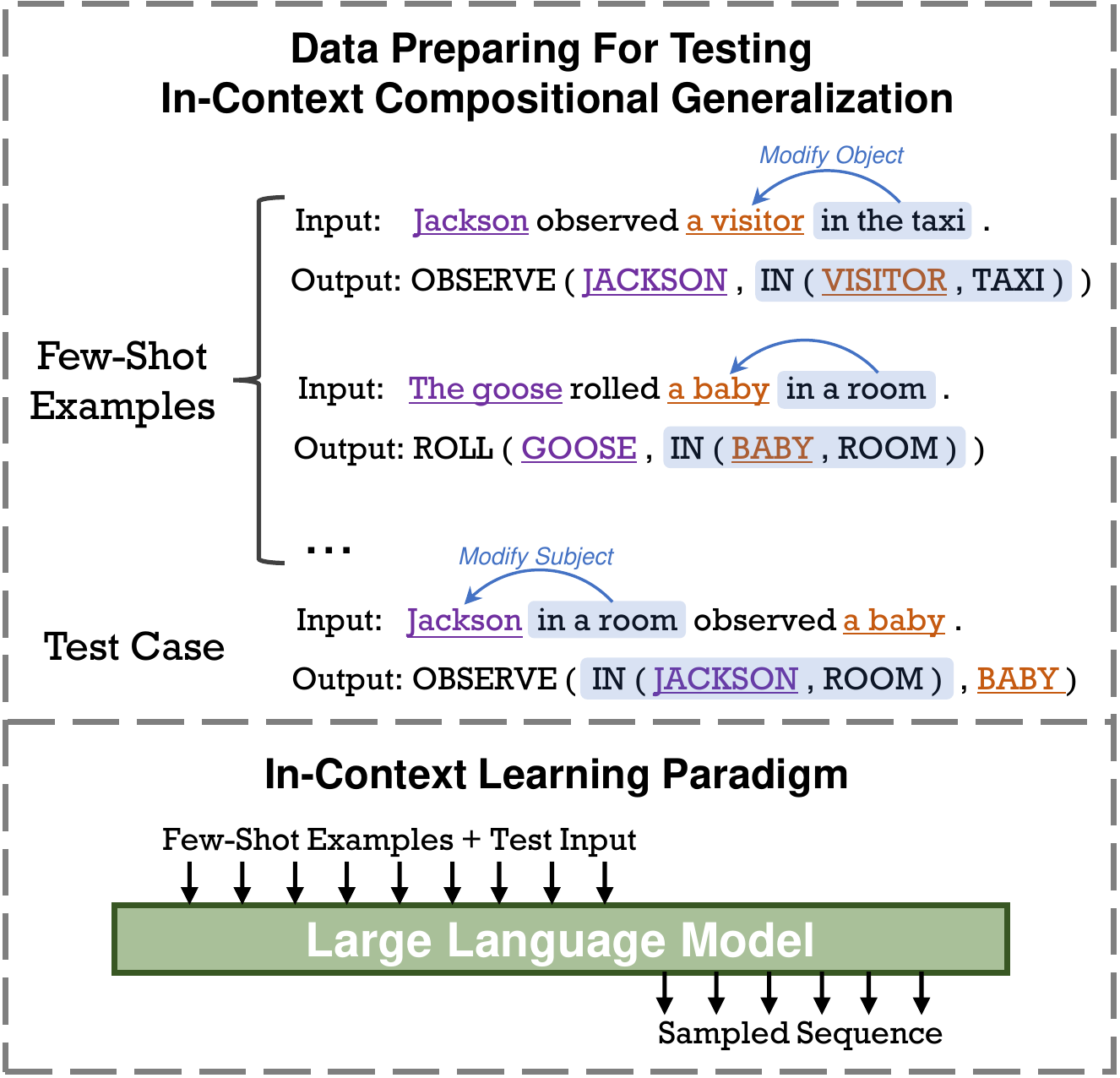}
    \caption{
    Test compositional generalization under in-context learning.
    This case belongs to \textit{Phrase Recombination} in \textsc{CoFe}.
    The \fcolorbox[HTML]{DBE3F4}[HTML]{DBE3F4}{phrases} modify the \textcolor[HTML]{C55A11}{\underline{objects}} in examples  but are recombined with \textcolor[HTML]{7030A0}{\underline{subject}} in test input.
    }
    \label{fig:first_figure}
\end{figure}

Compositional generalization is an essential capability of human intelligence.
It means to understanding and producing novel expressions by recombining known components in language~\cite{chomsky1957syntactic, montague1974english, fodor2002compositionality}.
Taking examples in Figure~\ref{fig:first_figure}, after learning the combination ``\textit{baby in a room}'', human intelligence can easily generalize to ``\textit{Jackson in a room}''.
On exploring this human-like capability in deep learning models, several benchmarks such as SCAN~\citep{lake2018generalization}, CFQ~\citep{keysers2019measuring} and COGS~\citep{kim2020cogs} have been proposed based on semantic parsing\footnote{Semantic parsing means translating natural language (NL) expressions into semantic representations (i.e., logical forms).} tasks.
In these benchmarks, the training set cover all the primitives while lacking certain combinations, and the test set focuses on these missing combinations.
By fine-tuning generic neural models on these benchmarks, much work reported that these models exhibit poor compositional generalization~\citep{furrer2020compositional, shaw2021compositional, bogin2022unobserved}.

Recently, in-context learning with large language models exhibits impressive performance on various tasks~\citep{brown2020language, rae2021scaling, wei2022emergent}.
By conditioning on few-shot in-context examples, the pre-trained language model, with extremely large model size and pre-trained corpus, can perform downstream tasks without any update on pre-trained parameters.

Behind the impressive performance of in-context learning, we are curious whether this prevailing paradigm can take a step towards compositional generalization.
To investigate this, we first take an initial exploration: for each test case in COGS, we select in-context examples from its training set and ensure that all primitives in each test case are covered by the equipped in-context examples.
Our initial exploration suggests that compositional generalization can be easily affected by in-context examples:
with only covering primitives, davinci 175B lags behind fine-tuned GPT2-Large with 24.2\% accuracy (similar to the observation in \citet{qiu2022evaluating});
with also covering some local structures (inspired by \citet{bogin2022unobserved}), davinci outperforms fine-tuned GPT2-Large with 3.9\% accuracy.
Based on these initial observations, we raise and investigate the question: \textit{How do in-context examples affect compositional generalization?}

We construct the test suite \textsc{CoFe} (based on COGS) to facilitate our systematic investigation.
Taking the coverage of primitives as a basic principle in \textsc{CoFe}, we further define and inject three factors in selecting in-context examples: similarity, diversity, and complexity.
Similarity is considered as the matching of hidden structures behind concrete expressions.
Diversity reflects whether the context presents repeated patterns or not.
Complexity portrays the amount of information contained in each example.
By controlling these factors in constructing \textsc{CoFe}, we can systematically investigate how would in-context examples influence the performance on compositional generalization.

Our experiments demonstrate that all three factors matter for in-context compositional generalization.
We leverage six large language models in GPT series: davinci, 
code-cushman-001, code-cushman-002, 
text-davinci-002, text-chat-davinci-002, and code-davinci-002.
The observations are consistent across models:
to better perform compositional generalization, all backbone models prefer in-context examples with higher structural similarity to the test case, higher diversity among different examples, and lower complexity in each individual example.
Furthermore, beyond the influence from these factors, in-context compositional generalization still faces two challenges.
One is that in-context learning has difficulty recombining fictional words (e.g., random tokens) rather than commonly used ones.
The other one is that in-context examples are still required to cover the linguistic structures in NL expressions, even though the backbone model has been pre-trained on large corpus.

Our contributions are three-fold:
1) to answer the research question posed, we investigate three factors in selecting in-context examples and draw consistent conclusions across models;
2) we construct \textsc{CoFe} to conduct our systematic investigation, and will release it to facilitate further exploration of in-context compositional generalization;
3) we also point out two remaining challenges that in-context learning still struggles to handle.
We hope our analysis would provide insights on how to select proper in-context examples,
and to shed light on the future research of in-context compositional generalization.
\textsc{CoFe} is publicly available at \url{https://github.com/microsoft/ContextualSP/tree/master/cofe}.

\begin{figure*}[t]
    \centering
    \includegraphics[width=.95\textwidth]{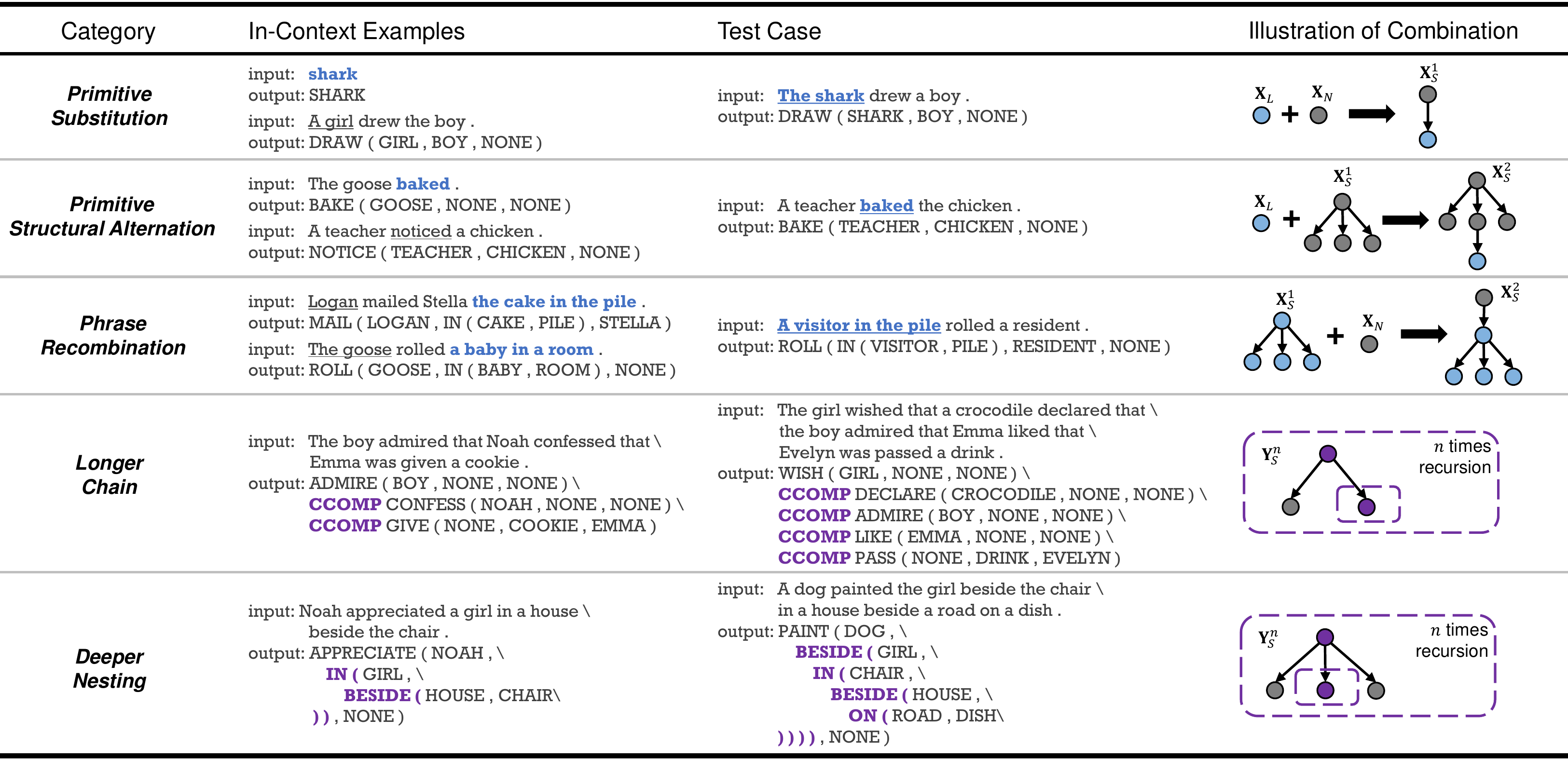}
    \caption{
    Five categories of aiming combinations.
    The key parts in combinations are marked with \underline{underlines} and colors (\textcolor[HTML]{4472C4}{blue} in NL-side and \textcolor[HTML]{7030A0}{purple} in code-side).
    The last column follows the notations defined in Section~\ref{sec:formalize}.
    }
    \label{fig:category}
\end{figure*}

\section{In-Context Compositional Generalization}\label{sec:in_context_cg}
In-context compositional generalization refers to understand and produce novel combinations through recombining the building blocks presented by in-context examples.
We first  introduce some basic settings for testing this desired capability, then show our initial observations.

\subsection{Principles for Measuring In-Context Compositional Generalization}

To measure in-context compositional generalization under a test suite, each test case and its equipped in-context examples should satisfy two principles.
\begin{itemize}

    \item \textbf{Combination held-out principle}: to test generalization on certain combinations, in-context examples should exclude these combinations while test cases contain them.

    \item \textbf{Primitive coverage principle}: the \textbf{primitives} contained in each test case should be fully covered by in-context examples.
    Primitives are the minimum indivisible units in expressions.
    In this work, we mainly consider primitives as \textbf{lexical items} (e.g., the noun ``\textit{baby}'' and the verb ``\textit{observed}'' in Figure~\ref{fig:first_figure}).
    
\end{itemize}
We say that a model exhibits in-context compositional generalization if it performs well on a test suite that satisfies these two principles.

\subsection{COGS (Under In-Context Learning)}\label{sec:cogs}

COGS is a compositional generalization benchmark designed for the fine-tuning paradigm:
based on a semantic parsing task, the training set of COGS covers all primitives in this task, while several combinations of primitives in the test set are excluded from the training set.
We term these excluded combinations as \textbf{aiming combinations}.

We measure in-context compositional generalization based on COGS, by converting it from the original fine-tuning paradigm to the in-context learning paradigm.
For each COGS test case, we select in-context examples from the training set $\mathcal{B}$, ensuring that the two principles are satisfied.
Note that, for each test case, there are usually different collections of in-context examples satisfying the two principles.
Our basic setting is to use a random one among them, and we show that this casual strategy could lead to an underestimation of in-context compositional generalization (Section~\ref{sec:initial_explore}).

To facilitate testing on more complex logical forms, we reconstruct some target-side clauses from the chain structure into the nested-function format (illustrated in Figure~\ref{fig:category}).
This reconstruction follows \citet{an2023does} and is similar to the conversion from Lambda calculus to FunQL in Geo domain\citep{zelle1996learning, kate2005learning, zettlemoyer2012learning}.
Moreover, to improve human readability, we omitted two types of details:
the special marker for definite descriptions and the Skolem constants.
These details do not affect the testing of compositional generalization.
Apart from these omitted details, the logical forms in \textsc{CoFe} unambiguously represent the main semantics in the domain of COGS, such as semantic roles, modifications, and orders among clauses and modifications.
More details about \textsc{CoFe} logical forms are contained in Appendix~\ref{sec:ap_grammar}.

\paragraph{Categories of aiming combinations.}
The aiming combinations in COGS can be divided into five categories, of which two are \textbf{low-level combinations} (i.e., focusing on specific primitives) and three are \textbf{high-level combinations} (i.e., focusing on high-level structures), illustrated in Figure~\ref{fig:category}.

\begin{itemize}
    \vspace{-0.5em}
    \item Primitive Substitution (\textit{PrimSubs}): Compose a primitive (e.g., ``\textit{shark}'') with a grammatical role (e.g., ``\textit{subject}'').

    \vspace{-0.5em}
    \item Primitive Structural Alternation (\textit{PrimAlte}): Compose a primitive (e.g., ``\textit{baked}'') with a sentence structure (e.g., ``\textit{subj. verb obj.}'').

    \vspace{-0.5em}
    \item Phrase Recombination (\textit{PhraReco}): Compose a prepositional phrase (e.g., ``\textit{A in B}'') with a grammatical role (e.g., ``\textit{subject}'').

    \vspace{-0.5em}
    \item Longer Chain (\textit{LongChain}): 
    Extend the tail of the logical form with $\mathtt{CCOMP}$ clauses $\in\mathbf{Y}_{S}^{1}$.
    The max recursive times of $\mathtt{CCOMP}$ clauses in $\mathcal{B}$ is 2, while in test case it is 12.

    \vspace{-0.5em}
    \item Deeper Nesting (\textit{DeepNest}):
    Expand the arguments in functions with $\mathtt{IN/ON/BESIDE}$ clauses $\in\mathbf{Y}_{S}^{1}$.
    The max recursive times in $\mathcal{B}$ and test cases are the same with \textit{LongChain}.
    
\end{itemize}
\vspace{-0.5em}
\noindent Note that \textit{PrimSubs} and \textit{PrimAlte} are low-level combinations while others are high-level ones.

\subsection{In-Context Learning vs Fine-Tuning}\label{sec:initial_explore}

Compositional generalization under the fine-tuning paradigm has been widely studied~\citep{furrer2020compositional, shaw2021compositional, bogin2022unobserved}, while there is little observation under in-context learning.
To first get a general sense about in-context compositional generalization, we conduct an initial exploration to compare with a fine-tuning baseline.

\paragraph{Models and setups.}
We test in-context compositional generalization with six large models in GPT series: 
davinci, 
code-cushman-001 (cuchman001), code-cushman-002 (cuchman002), 
text-davinci-002 (text002), text-chat-davinci-002 (chat002), and code-davinci-002 (code002).
The sampling temperature is 0 (i.e., greedy decoding), and the max decoding length is 500.
The reported metric is exact-match accuracy.
To set a fine-tuning baseline, we take GPT2-Large with 0.7B parameters.
We fine-tune it on the whole $\mathcal{B}$ and test without in-context examples.
We set learning rate as 1e-5 and batch size as 8 during fine-tuning, and set beam size as 5 for inference.
Appendix~\ref{sec:ap_models} includes more details.

\begin{figure}[t]
    \centering
    \includegraphics[width=.47\textwidth]{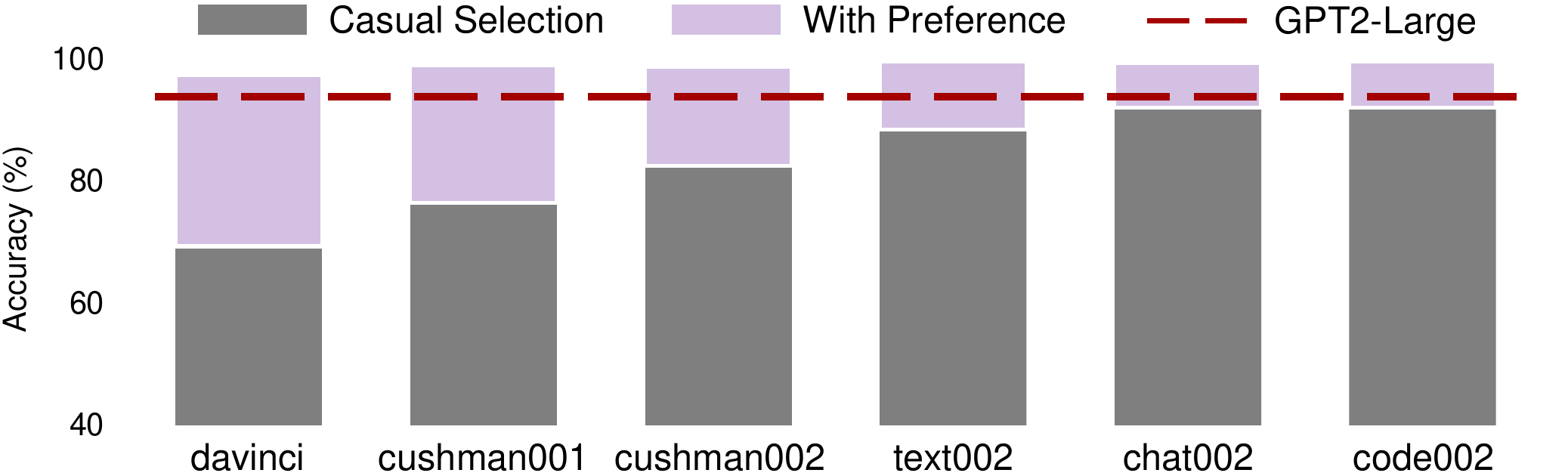}
    \caption{Initial observations on \textit{PrimSubs}: casual selection leads to low performance  while adding preference brings considerable gains.
    }
    \label{fig:initial}
\end{figure}

\paragraph{Casual selection leads to low performance of in-context compositional generalization.}
For selecting in-context examples, we first take a \textit{casual selection}: while satisfying the primitive coverage principle, we randomly select 10 examples without other preference.
We conduct initial exploration on \textit{PrimSubs} category.
Figure~\ref{fig:initial} shows that under the casual selection, all six models lag behind the fine-tuned GPT2-Large on \textit{PrimSubs}.
In particular, although the size of davinci is more than 200 times that of GPT2-Large, there is a 24.2\% accuracy gap between davinci and the fine-tuned GPT2-Large.
These observations are close to \citet{qiu2022evaluating}.

However, we suppose the potential of in-context learning is still not fully revealed.
Specifically, the selection of in-context examples does not yet take full advantage of available examples in $\mathcal{B}$.
In next try, while still following the primitive coverage principle, we consider injecting some additional preference in the selection of in-context examples.

\paragraph{Preference in selection could bring huge improvement on \textit{PrimSubs}.}
Inspired by \citet{bogin2022unobserved} that suggests the influence of unobserved local structures, we consider to prioritize examples that have similar hidden structures to the test case.
Figure~\ref{fig:initial} shows that with this preference in selection, results on \textit{PrimSubs} hugely change: davinci now outperforms the fine-tuned GPT2-Large;
code-davinci-002 even performs near-perfectly.
These changes strongly suggest that the selection of in-context examples can significantly affect in-context compositional generalization.

Based on these initial results, to further reveal the potential of in-context learning, we perform in-depth investigations on how the selection of in-context examples affects compositional generalization.

\section{Factors Under In-Context Examples}\label{sec:factors}

To facilitate our systematic investigation, we construct \textbf{\textsc{CoFe}} (\textbf{\textsc{Co}}mpositional generalization with \textbf{\textsc{Fe}}w-shot examples), which is derived from COGS.
For selecting in-context examples in constructing \textsc{CoFe}, we identify, inject, and control three potential factors: similarity, diversity, and complexity.

\subsection{Conceptual Definitions}\label{sec:conceptual_definition}

We first give conceptual definitions of our considered factors and discuss our intuitions behind them.

\paragraph{Similarity}
has been widely considered as the main factor in selecting in-context examples~\citep{liu2022makes, shin2021constrained, rubin2021learning, poesia2021synchromesh}.
The primitive coverage principle can be regarded as a basic \textit{lexical similarity} on the surface of expressions.
Beyond this surface similarity, we consider that the \textbf{structural similarity} hidden behind expressions could be a beneficial factor.
From the view of syntactic structure, the recombination of primitives is equivalent to the reconstruction of the parse tree.
Similar structures would ease the difficulty of recombination because the model does not need to completely reconstruct the entire  structure of in-context examples.
Moreover, some work has suggested that the challenge of compositional generalization under fine-tuning lies in unobserved structures~\citep{keysers2019measuring, shaw2021compositional, bogin2022unobserved}.

\paragraph{Diversity}
concerns the \textbf{repetitiveness} among in-context examples.
It portrays the property among in-context examples.
Specifically, the context is under low diversity if it contains many repeating patterns among in-context examples, otherwise it is under high diversity.
Under in-context learning, the low diversity can easily lead to biased observations on the full task space, as there are only few examples for the model to learn.
Thus, we suppose that the low diversity among examples could block in-context compositional generalization.
Moreover, some work also demonstrated that the diversity in training data could affect compositional generalization under fine-tuning~\citep{oren2021finding}.

\paragraph{Complexity}
reflects the amount of information contained in each individual in-context example.
The higher complexity means that the example could provide more information to the model, but these information could be redundant.
In addition, the difficulty in directly learning from complex examples has been flagged at the intersection of cognitive science and machine learning~\citep{Elman1993LearningAD, bengio2009curriculum}.
Such difficulty may be more severe for in-context learning, since the parameters of the model cannot be updated to fit these complex examples.
Thus, we suppose that too high complexity might hinder performance.

\subsection{Incorporate Three Factors Into Test Suite}\label{sec:formalize}

To inject these factors in selecting in-context examples, we design a \textit{matching score} based on the parse trees behind concrete expressions.
Formally, considering the primitive coverage, \textcolor[HTML]{02882F}{structural similarity}, \textcolor[HTML]{8409E9}{diversity} and \textcolor[HTML]{C00000}{complexity}, the matching score of two parse trees $\mathbf{T}$ and $\mathbf{T'}$ is defined as follows,
\begin{equation}\label{equ:match}
\begin{split}
    \mathtt{Match}(\mathbf{T}, \mathbf{T'}) = 
    & w_p\cdot|\mathtt{P}(\mathbf{T})\cap\mathtt{P}(\mathbf{T'})| + \\
    & \textcolor[HTML]{02882F}{w_s\cdot |\mathtt{S}(\mathbf{T})\cap \big[ \mathtt{S}(\mathbf{T'})} \textcolor[HTML]{8409E9}{-\mathtt{S}(\mathcal{C})} \textcolor[HTML]{02882F}{\big] |} - \\
    & \textcolor[HTML]{C00000}{w_c\cdot\mathtt{depth}(\mathbf{T'})},
\end{split}
\end{equation}

\vspace{-2mm}
\noindent in which $\mathtt{P}(\cdot)$ contains primitives, $\mathtt{S}(\cdot)$ contains \textit{partial structures} (defined later), $\mathcal{C}$ contains already selected examples, $\mathtt{S}(\mathbf{T'})-\mathtt{S}(\mathcal{C})$ means to exclude already covered parts in $\mathtt{S}(\mathcal{C})$ from $\mathtt{S}(\mathbf{T'})$,
and $\mathtt{depth}(\cdot)$ reflects the complexity of the tree.

The meaning of three factors in Equation~\ref{equ:match} is that:
the structural similarity means covering $\mathtt{S}(\mathbf{T})$, the high diversity means to avoid repeatedly covering the same element in $\mathtt{S}(\mathbf{T})$, and the low complexity is to prioritize low-depth structures.

Based on this matching score, the overall ranking score between the test case $(\mathbf{X}, \mathbf{Y})$ and a candidate $(\mathbf{X}_{c}, \mathbf{Y}_{c})$ is calculated as follows,
\begin{equation}\label{equ:score}
\mathtt{score}_c = \mathtt{Match}(\mathbf{X}, \mathbf{X}_{c}) + \mathtt{Match}(\mathbf{Y}, \mathbf{Y}_{c}),
\end{equation}
in which both the matching of source side (i.e., NL expressions) and target side (i.e., logical forms) are considered.
\citet{poesia2021synchromesh} has demonstrated the importance of target-side similarity in semantic parsing and code generation tasks, and this work will further investigates the necessity of source-side matching.
In the following, we will give a more detailed description of notations in Equation~\ref{equ:match}.

\begin{figure}[t]
    \centering
    \includegraphics[width=.47\textwidth]{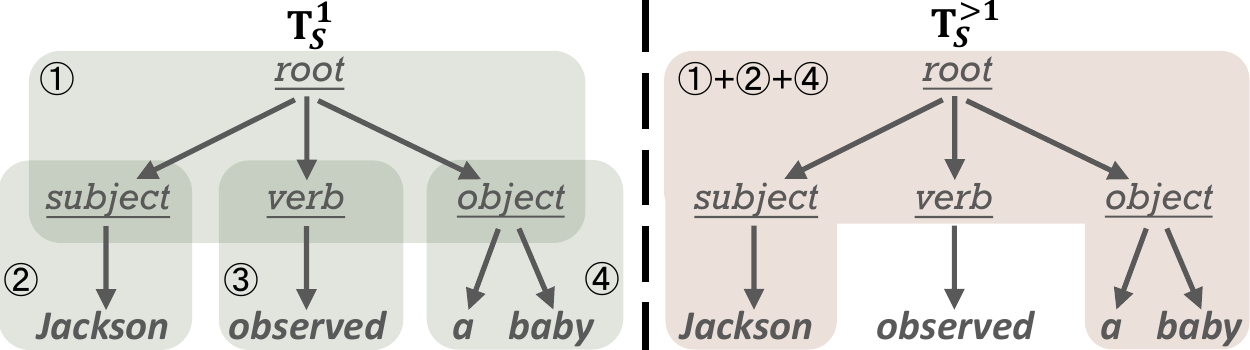}
    \caption{
    $\mathbf{T}_{S}^{1}$ and $\mathbf{T}_{S}^{>1}$ in the parse tree of the expression ``\textit{Jackson observed a baby}''.
    $\mathbf{T}_{S}^{1}$ contains four \fcolorbox[HTML]{E1E6DE}[HTML]{E1E6DE}{one-depth sub-structures}.
    We only illustrate one \fcolorbox[HTML]{ECE2DB}[HTML]{ECE2DB}{combination} $\in\mathbf{T}_{S}^{>1}$ composed from \fcolorbox[HTML]{E1E6DE}[HTML]{E1E6DE}{\ding{172}}, \fcolorbox[HTML]{E1E6DE}[HTML]{E1E6DE}{\ding{173}} and \fcolorbox[HTML]{E1E6DE}[HTML]{E1E6DE}{\ding{175}} $\in\mathbf{T}_{S}^{1}$.
    $\mathbf{T}_{L}$ are in \textbf{bold} and $\mathbf{T}_{N}$ are with \underline{underlines}.
    }
    \label{fig:substructure}
\end{figure}

\paragraph{Detailed description:}
Figure~\ref{fig:substructure} shows an illustration of notations.
Considering an expression $e$ with the parse tree $\mathbf{T}$,
$\mathbf{T}_{L}$ represents leaf nodes (e.g., ``\textit{Jackson}'') and $\mathbf{T}_{N}$ contains internal nodes (e.g., ``\textit{subject}'').
$\mathbf{T}_{S}^{1}$ contains one-depth sub-structures in $\mathbf{T}$.
Each $\mathbf{T}_{s}^{1} \in \mathbf{T}_{S}^{1}$ (e.g., \ding{172} in Figure~\ref{fig:substructure}) contains one parent node (e.g., ``\textit{root}'') and a set of child nodes (e.g., ``\textit{subject}'', ``\textit{verb}'' and ``\textit{object}'').
$\mathbf{T}_{S}^{>1}$ contains deeper sub-structures that are composed from several one-depth sub-structures in $\mathbf{T}_{S}^{1}$ (e.g., \ding{172}+\ding{173}+\ding{175} in Figure~\ref{fig:substructure}).
In Equation~\ref{equ:match}, the primitives $\mathtt{P}(\mathbf{T})=\mathbf{T}_{L}$, and the partial structures $\mathtt{S}(\mathbf{T})=\mathbf{T}_{S}^{1}\cup\mathbf{T}_{S}^{>1}$.
Note that aiming combinations $\subset \mathtt{S}(\mathbf{T})$.
Appendix~\ref{sec:ap_notations} includes more details.

\begin{table*}[t]
\renewcommand\arraystretch{1.2}
\Huge
\caption{Results with (and without) structural similarity.
\fcolorbox{gray!25}{gray!25}{Grey boxes} mark the significantly better performances compared to the fine-tuned GPT2-Large.
}
\label{tab:structural_similarity}
\centering
\resizebox{.9\linewidth}{!}{
\begin{tabular}{@{}p{8cm}<{\centering}p{9.5cm}<{}p{4.5cm}<{\centering}p{4.5cm}<{\centering}p{4.5cm}<{\centering}p{4.5cm}<{\centering}p{4.5cm}<{\centering}p{4.5cm}<{\centering}@{}}
\toprule
Model & \multicolumn{1}{c}{Setting} & PrimSubs & PrimAlte & PhraReco & LongChain & DeepNest & Avg. Acc \\ \midrule
\multirow{2}{*}{code-davinci-002} & \enspace Primitive Coverage & 92.2 & 77.1 & 60.8 & 62.1 & 12.3 & 60.9 \\
 & \enspace  + Structural Similarity & 
 \cellcolor{gray!25}99.8 & \cellcolor{gray!25}99.7 & \cellcolor{gray!25}65.3 & \cellcolor{gray!25}87.0 & 26.0 & \cellcolor{gray!25}75.6 \\ \midrule
\multirow{2}{*}{text-chat-davinci-002} & \enspace Primitive Coverage & 92.2 & 75.4 & 47.0 & 65.0 & 6.3 & 57.2 \\
 & \enspace  + Structural Similarity & \cellcolor{gray!25}99.5 & \cellcolor{gray!25}99.3 & 53.4 & \cellcolor{gray!25}87.7 & 18.9 & \cellcolor{gray!25}71.8 \\ \midrule
\multirow{2}{*}{text-davinci-002} & \enspace Primitive Coverage & 88.5 & 66.4 & 38.7 & 46.5 & 2.9 & 48.6 \\
 & \enspace  + Structural Similarity & \cellcolor{gray!25}99.7 & \cellcolor{gray!25}99.4 & 39.4 & \cellcolor{gray!25}80.2 & 12.7 & \cellcolor{gray!25}66.3 \\ \midrule
\multirow{2}{*}{code-cushman-002} & \enspace Primitive Coverage & 82.6 & 55.6 & 21.3 & 29.3 & 5.0 & 38.8 \\
 & \enspace  + Structural Similarity & \cellcolor{gray!25}98.9 & \cellcolor{gray!25}99.0 & 28.5 & 64.0 & 15.1 & 61.1 \\ \midrule
\multirow{2}{*}{code-cushman-001} & \enspace Primitive Coverage & 76.6 & 60.7 & 16.9 & 5.0 & 1.0 & 32.0 \\
 & \enspace  + Structural Similarity & \cellcolor{gray!25}99.1 & 98.4 & 20.7 & 11.1 & 8.9 & 47.6 \\ \midrule
\multirow{2}{*}{davinci} & \enspace Primitive Coverage & 69.4 & 52.3 & 9.4 & 2.3 & 0.2 & 26.7 \\
 & \enspace  + Structural Similarity & \cellcolor{gray!25}97.5 & 95.4 & 12.3 & 13.4 & 1.4 & 44.0 \\ \midrule\midrule
Fine-Tuning Baseline & \multicolumn{1}{c}{-} & 93.6 & 97.9 & 14.0 & 5.4 & 0.0 & 42.2 \\ \bottomrule
\end{tabular}
}
\end{table*}

\section{Experiments and Analysis}\label{sec:experiments_and_analysis}

\subsection{Experimental Settings and Hyper-Parameters}
We take a greedy-search algorithm to sequentially select 10 examples for each test case.
Models and setups follow our initial explorations in Section~\ref{sec:initial_explore}.
For the investigation of each factor, hyper-parameters in Equation~\ref{equ:match} are set as follows\footnote{Appendix~\ref{sec:ap_implement} contains our detailed implementations.}.

In all settings, we prioritize the matching of primitives (i.e., $|\mathtt{P}(\mathbf{T})\cap\mathtt{P}(\mathbf{T'})|$ in Equation~\ref{equ:match}) since the primitive coverage principle should be firstly satisfied.
Concretely, we set $w_p=100$ and ensure $w_p\gg w_s$ and $w_c$ in all settings.

For investigating structural similarity\footnote{This setting is named \textit{full similarity setting}.}, we set $w_s=1$ and $w_c=0$, and exclude $\mathtt{S}(\mathcal{C})$ term.

For investigating the effect of higher diversity, we add the $\mathtt{S}(\mathcal{C})$ term and keep other settings.

For complexity, we set $|w_c|\cdot \mathrm{max}(\mathrm{depth}(\mathbf{T'})) < w_s$, such that the of preference of complexity will not influence the priority of structural similarity.
Concretely, as $\mathrm{max}(\mathrm{depth}(\mathbf{T'}))=12$ in \textsc{CoFe}, we set $w_c=0.01$ for the low-complexity experiments and $w_c=-0.01$ for the high-complexity experiments, and exclude $\mathtt{S}(\mathcal{C})$ term.

Some basic statistics for \textsc{CoFe} under full similarity setting are listed in Table~\ref{tab:statistics}, and Appendix~\ref{sec:statistics_diversity_complexity} contains statistics under other settings.
These statics show that the primitive coverage principle is well satisfied, since the cover rates of $\mathbf{T}_{L}$ are almost 100\%.
Note that the coverage on $\mathbf{T}_{S}^{1}\cup\mathbf{T}_{S}^{>1}$ must be lower than 100\% since the aiming combination must be excluded.

\begin{table*}[t]
\caption{Basic statistics of \textsc{CoFe}.
}
\label{tab:statistics}
\centering
\resizebox{.8\linewidth}{!}{
\begin{tabular}{@{}l|c|cccc|ccc@{}}
\toprule
\multirow{2}{*}{Statistics} & \multirow{2}{*}{Number of Instances} & \multicolumn{4}{c|}{Average Coverage} & \multicolumn{3}{c}{Average Length} \\
 &  & $\mathbf{T}_{L}$ & $\mathbf{T}_{N}$ & $\mathbf{T}_{S}^{1}$ & $\mathbf{T}_{S}^{>1}$ & Context & Case Input & Case Output \\ \midrule
Test Cases & 4,785 & 99.7\% & 100\% & 88.9\% & 49.3\% & 297.7 & 17.8 & 33.7 \\
\quad - PrimSubs & 1,100 & 100\% & 100\% & 79.8\% & 45.1\% & 236.7 & 7.1 & 11.5 \\
\quad - PrimAlte & 700 & 100\% & 100\% & 96.6\% & 59.7\% & 269.4 & 7.9 & 13.8 \\
\quad - PhraReco & 1,000 & 100\% & 100\% & 84.4\% & 19.8\% & 254.0 & 10.7 & 16.9 \\
\quad - LongChain & 1,000 & 99.8\% & 100\% & 97.8\% & 76.7\% & 370.6 & 32.4 & 76.7 \\
\quad - DeepNest & 985 & 98.8\% & 100\% & 89.0\% & 48.6\% & 356.4 & 29.0 & 46.3 \\ \midrule
Example Bank & 24,155 & - & - & - & - & - & 7.5 & 10.5 \\ \bottomrule
\end{tabular}
}
\end{table*}

\subsection{Similarity}\label{sec:similarity}

\paragraph{Structural similarity brings significant gains.}
Table~\ref{tab:structural_similarity} shows the performance with structural similarity.
Compared to the results without structural similarity (i.e., only with the coverage on primitives), there are considerable gains on all five categories and across all six models.
These gains clearly demonstrate that beyond primitive coverage, the structural similarity under in-context examples are essential for compositional generalization.

\begin{figure*}[t]
    \centering
    \includegraphics[width=.99\textwidth]{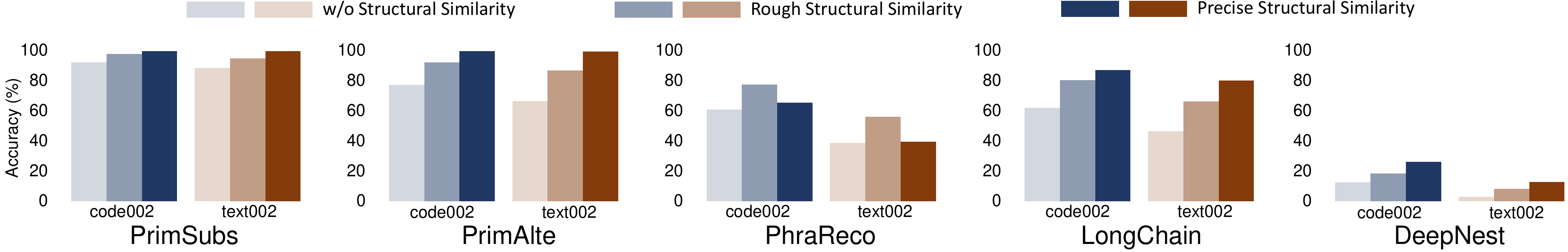}
    \caption{
    Performance of code-davinci-002 and text-davinci-002 with different levels of structural similarity.
    }
    \label{fig:precise_structural_similarity}
\end{figure*}

\paragraph{More precise structural similarity brings larger gains.}
As mentioned in Section~\ref{sec:formalize}, the structural similarity considers to match $\mathtt{S}(\mathbf{T})$ which contains two parts, $\mathbf{T}_{S}^{1}$ and $\mathbf{T}_{S}^{>1}$.
Specifically, we regard that $\mathbf{T}_{S}^{1}$ describes the \textit{rough structure} of $\mathbf{T}$, and $\mathbf{T}_{S}^{>1}$ determines a more \textit{precise structure}.
Based on the results in Table~\ref{tab:structural_similarity}, we are curious about whether a rough structural similarity is enough.
To verify this, we remove $\mathbf{T}_{S}^{>1}$ from $\mathtt{S}(\mathbf{T})$, which means that now we do not restrict the selected in-context examples to match precise structures in test cases.
Figure~\ref{fig:precise_structural_similarity} shows that the performances on four categories significantly drop with only a rough structural similarity, indicating that matching the precise structure of test case is still required for in-context examples.
The only exception lies in \textit{PhraReco}.
It suggests that similarity is not the only influential factor for in-context compositional generalization.
In Section~\ref{sec:diversity_complexity}, we will show that the low diversity and high complexity potentially cause this exception.

\paragraph{With structural similarity, low-level combinations are almost solved while high-level combinations still have large room for improvement.}
Specifically, for code-davinci-002, which exhibits the best performance among all backbone models, it performs near-perfectly on low-level combinations (i.e., \textit{PrimSubs} and \textit{PrimAlte}) while still does not achieve >95\% accuracy on high-level combinations  (i.e., \textit{PhraReco}, \textit{LongChain} and \textit{DeepNest}).
Although in-context learning greatly exceeds the fine-tuning baseline on high-level combinations, we suppose there is still potential for improvement.
Compared to low-level combinations, handling high-level ones requires more creation than imitation, thus just considering similarity for in-context examples is not enough.
In the following, we will further investigate these high-level combinations from the view of diversity and complexity.

\subsection{Diversity and Complexity}\label{sec:diversity_complexity}

\begin{figure}[t]
    \centering
    \includegraphics[width=.48\textwidth]{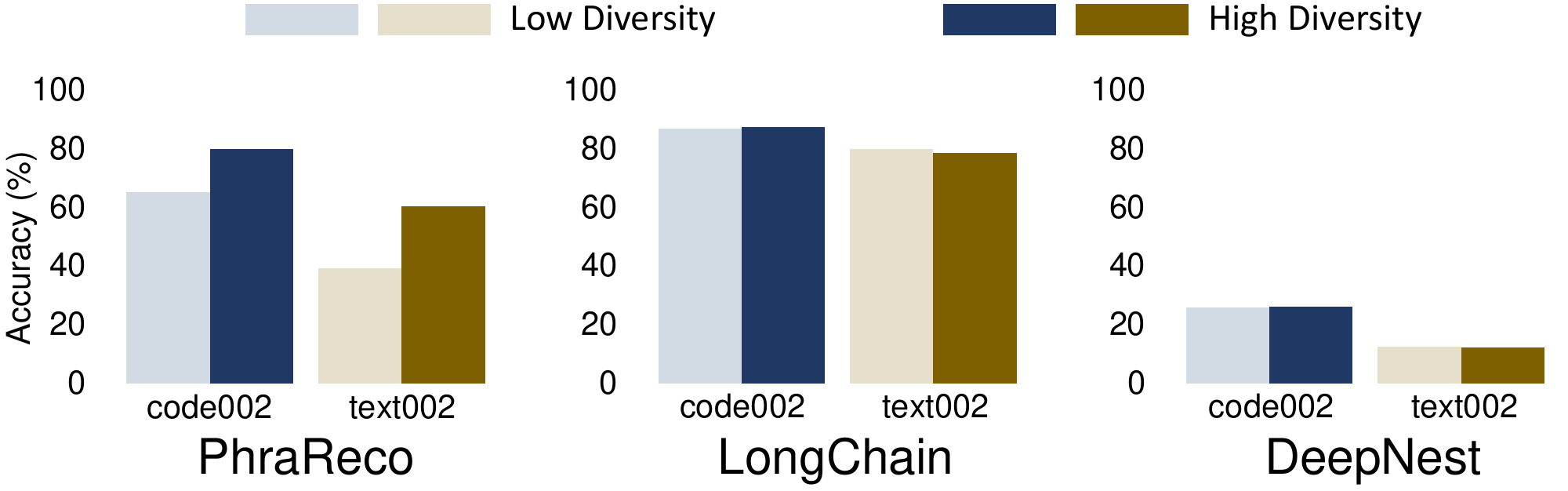}
    \caption{
    Performance under different diversity settings (on high-level combinations).
    }
    \label{fig:diversity}
\end{figure}

\paragraph{High diversity brings considerable gains on \textit{PhraReco}.}
Figure~\ref{fig:diversity} shows how diversity among in-context examples affects generalization on high-level combinations.
It shows that increasing the diversity could bring considerable gains in \textit{PhraReco}, while not affecting the other two categories.
For the performance on \textit{PhraReco}, the improvements from higher diversity are in line with our speculations in Section~\ref{sec:conceptual_definition}, that low diversity leads to biased observations, thus blocking high-level structural generalization.
For \textit{LongChain} and \textit{DeepNest}, 
beyond biased structures, their difficulty also lies in \textit{length generalization}, thus just increasing structural diversity brings less effect to them.

\begin{figure}[t]
    \centering
    \includegraphics[width=.48\textwidth]{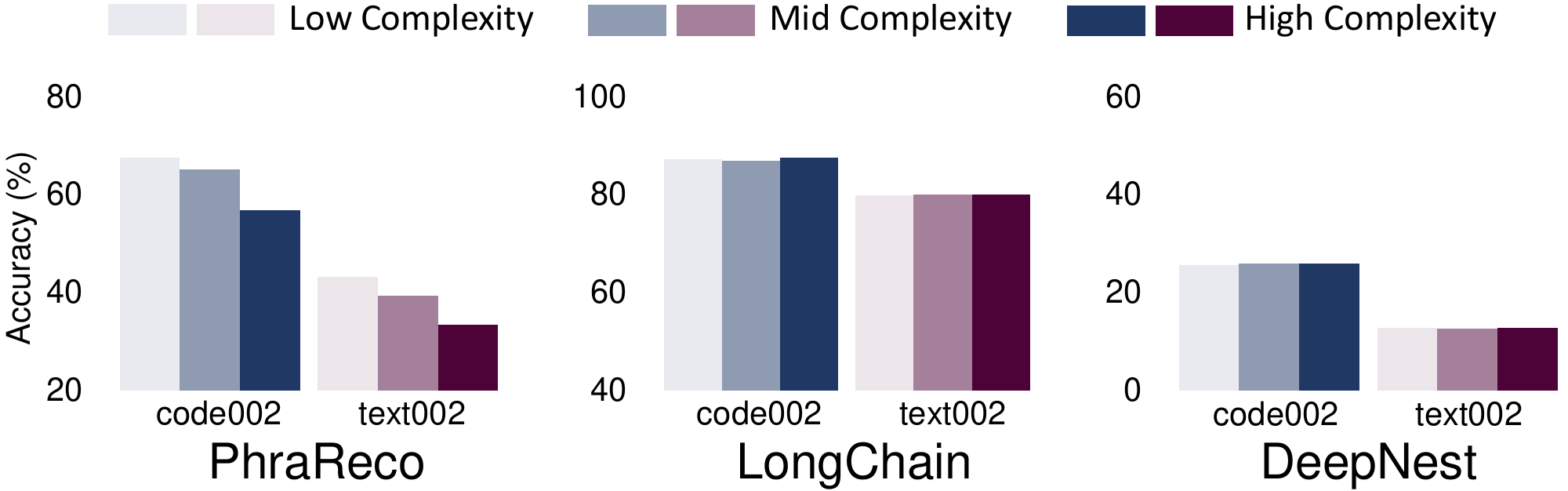}
    \caption{
    Performance under different complexity settings (on high-level combinations).
    }
    \label{fig:complexity}
\end{figure}

\paragraph{Low complexity brings considerable gains on \textit{PhraReco}.}
Figure~\ref{fig:complexity} shows how the complexity in each individual example affects generalization on high-level combinations.
For \textit{PhraReco}, there are $\sim$10\% gains in accuracy when the high complexity setting is changed to low complexity setting.
We suppose the reason behind this gain is that simple examples could reduce the learning difficulty for the model.
Moreover, simple examples also contain less redundant information thus would not confuse the model\footnote{Note that low and high complexity settings keep the same coverage rate on $\mathtt{S}(\mathbf{T})$, as demonstrated in Appendix~\ref{sec:statistics_diversity_complexity}.}.
For \textit{LongChain} and \textit{DeepNest}, there is still less change on performance.
Note that the max depth in these two categories is 13 while the max depth in the whole example bank is only 3.
Therefore, changing the complexity of in-context examples would bring negligible influence for test cases in \textit{LongChain} and \textit{DeepNest}.

\subsection{Analysis: Robustness to Prompt Order}\label{sec:robustness}

Some previous work on in-context learning showed that the order of exemplars in prompt could sometimes hugely influences the performance of LLMs~\citep{zhao2021calibrate, lu2022fantastically}.
Here, we examine whether our observations above are sensitive to the prompt order.
Based on the full similarity setting (Section~\ref{sec:similarity}), we consider three different strategies for ordering exemplars:
1) random order;
2) atom closer: exemplars with higher coverage on atomic blocks are placed closer to the test input;
3) structure closer (default): examples with higher similarity on linguistic structures are placed closer to the test input.
Implementations of different strategies for prompt order are detailed in Appendix~\ref{sec:ap_order}.

\begin{table}[t]
\renewcommand\arraystretch{1.2}
\Huge
\caption{Results under different prompt orders (full similarity setting).
$\Delta$ represents the max difference in performance for each model.
}
\label{tab:prompt_order}
\centering
\resizebox{.95\linewidth}{!}{
\begin{tabular}{@{}p{7cm}<{\centering}|ccc|p{2.5cm}<{\centering}@{}}
\toprule
Model & Structure Closer & Atom Closer & Random Order & $\Delta$ \\ \midrule
code-davinci-002 & \textbf{75.6} & 74.2 & 74.5 & 1.4 \\
text-davinci-002 & \textbf{66.3} & 66.0 & \textbf{66.3} & 0.3 \\
code-cushman-002 & \textbf{61.1} & 60.0 & 60.1 & 1.1 \\
code-cushman-001 & 47.6 & \textbf{48.2} & 47.3 & 0.9 \\
davinci & \textbf{44.0} & 43.6 & 42.5 & 1.5 \\ \bottomrule
\end{tabular}
}
\end{table}

Results in Table~\ref{tab:prompt_order} show that the performance only slightly changes under different prompt orders.
These results indicate that the main results revealed by \textsc{CoFe} is consistent and reliable.
It also indicates that in-context learning could be less sensitive to the prompt order when the in-context examples are chosen properly.

\subsection{Discussion: Difficulty in \textit{DeepNest}}

\begin{figure}[t]
    \centering
    \includegraphics[width=.48\textwidth]{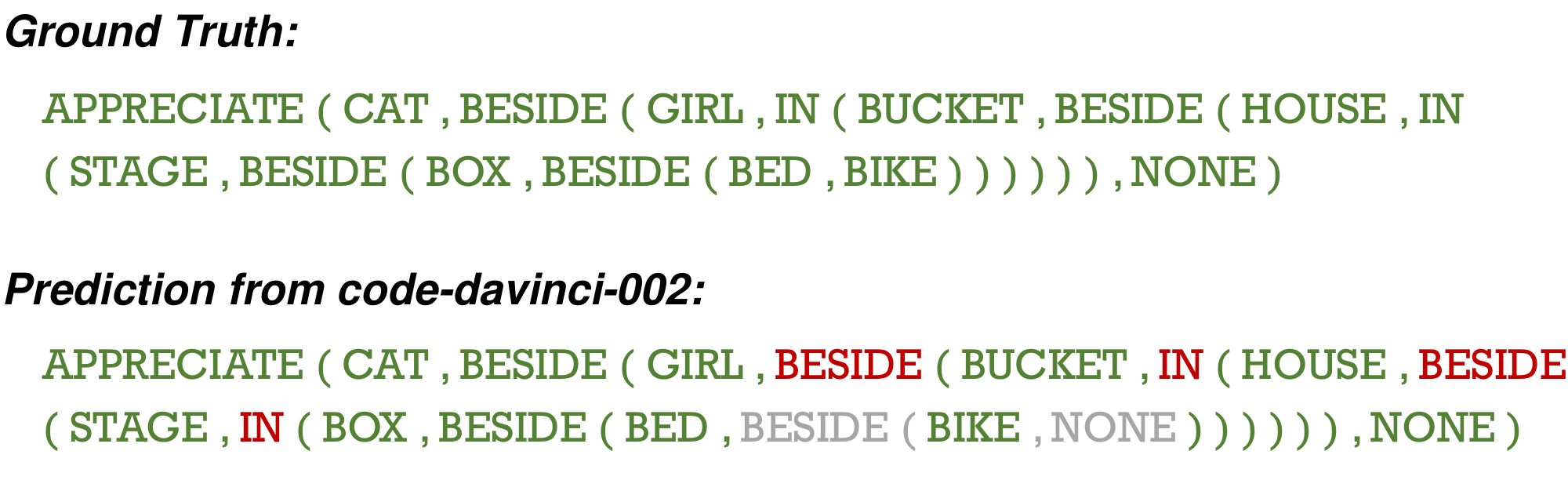}
    \caption{
    An error case in \textit{DeepNest} (full similarity setting) with \textcolor[HTML]{C00000}{wrong local words} and \textcolor[HTML]{A6A6A6}{redundant parts}.
    }
    \label{fig:word_mistakes}
\end{figure}

Among all five categories, in-context learning performs worst on \textit{DeepNest}.
Compared to \textit{LongChain} which also test recursive structures, the results on \textit{DeepNest} still lag far behind.
There is an interesting observation from the study of error cases (such as Figure~\ref{fig:word_mistakes}):
in-context learning frequently makes word-level mistakes, while the overall nested structure in the prediction is close to the ground truth.
It suggests that \textbf{the performance bottleneck in \textit{DeepNest} is to correctly fill the details in the complex structure}, rather than generating the sketch of the structure.
Appendix~\ref{sec:ap_case_study_deeper_nesting} provides further analysis.

\section{Remaining Challenges}\label{sec:challenges}

Our investigation has revealed a huge potential of in-context learning on performing compositional generalization\footnote{Appendix~\ref{sec:ap_full_results} shows the results of assembling factors.}.
Despite this potential, for achieving the ideal in-context compositional generalization, there remains the following two challenges.

\begin{figure*}[t]
    \centering
    \includegraphics[width=.99\textwidth]{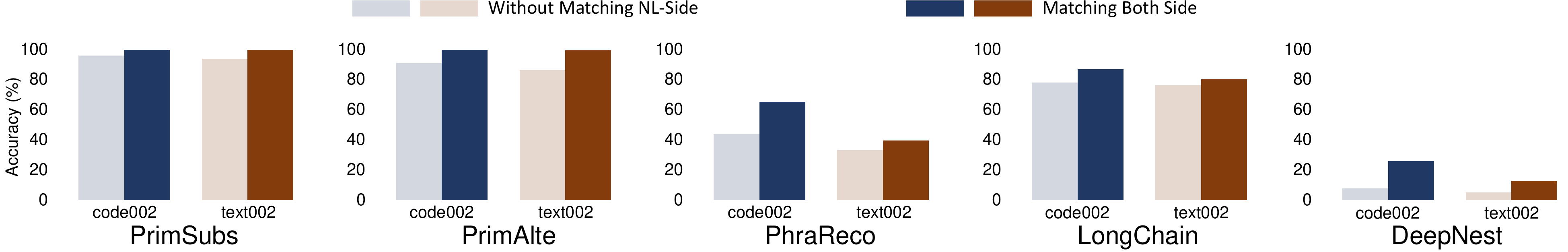}
    \caption{
    Performance with or without matching linguistic structures in NL expressions.
    }
    \label{fig:NL_similarity}
\end{figure*}

\begin{figure}[t]
    \centering
    \includegraphics[width=.48\textwidth]{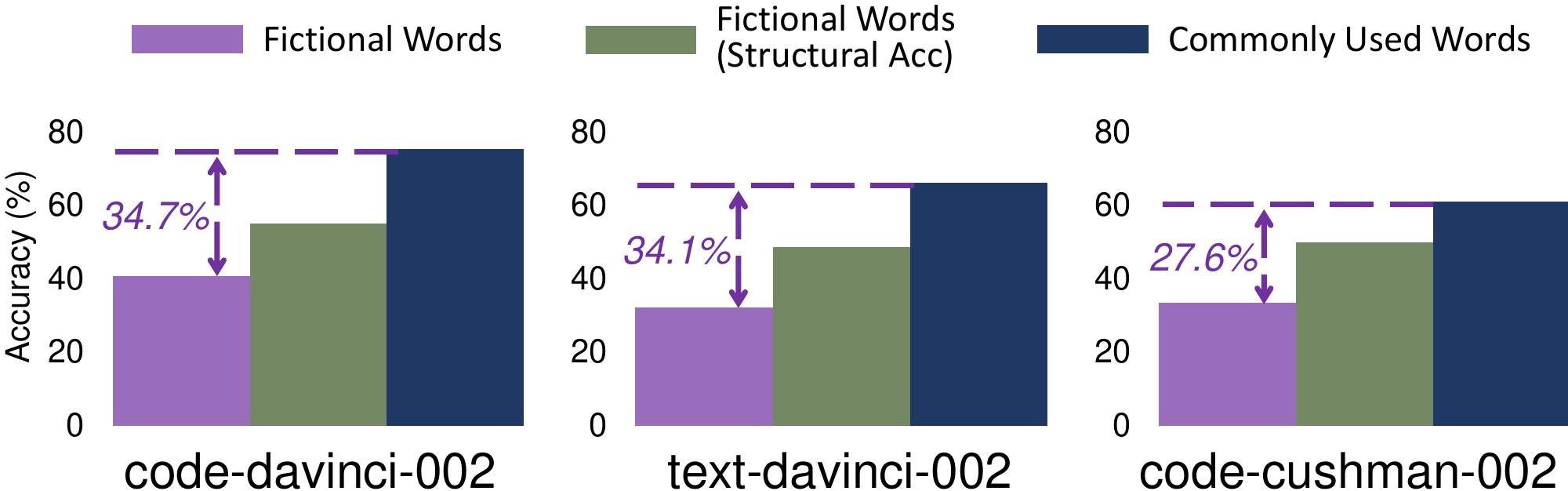}
    \caption{
    Average exact-match accuracy and structural accuracy with fictional words.
    }
    \label{fig:familiarity}
\end{figure}

\paragraph{In-context examples are still required to match linguistic structures in NL expressions.}
Since all backbone models have been pre-trained on large natural language corpus, we expect that these models could already handle the high variety in NL expressions without further hints from in-context examples.
Motivated by this, we conduct experiments on another variant of \textsc{CoFe}: the source-side term $\mathtt{Match}(\mathbf{X}, \mathbf{X}_{c})$ is removed from Equation~\ref{equ:score}, and the coverage of $\mathtt{S}(\mathbf{X})$ is limited (detailed in Appendix~\ref{sec:ap_NL_similarity}).
Figure~\ref{fig:NL_similarity} shows that on all five categories, the performance consistently drops if in-context examples do not match the NL-side structure.
It suggests that even having been pre-trained on large corpus, in-context learning still struggles to effectively recognize the semantic equivalence among different linguistic structures behind NL expressions (detailed in Appendix~\ref{sec:ap_case_study_semantic_equivalence}).

\paragraph{In-context learning has difficulty leveraging fictional words\footnote{The term ``fictional words'' means that these words are made up by us, so that large language models hardly encounter them during pre-training. Here, we generate fictional words by drawing random characters from the alphabet.}.}
The ideal compositional generalization requires that the recombination of primitives should be independent of the surface form in primitives.
In \textsc{CoFe}, we set the target-side primitives as the uppercase of source-side ones (e.g., ``\textit{cat}''$\rightarrow$``\textit{CAT}'').
Such \textit{case conversion} is commonly used in semantic parsing tasks.
To test whether in-context learning could use fictional words, we replace each target-side word with random characters (e.g., replace ``\textit{CAT}'' with ``\textit{MXR}'', detailed in Appendix~\ref{sec:ap_familiarity}).
Figure~\ref{fig:familiarity} shows the huge drops after changing words.
Moreover, we investigate the structural accuracy by only keeping the structural terminals (e.g., parentheses and commas) in predictions.
Figure~\ref{fig:familiarity} shows that the structural accuracy is also affected by fictional words.
It indicates that on performing in-context compositional generalization, the prediction of structural sketch is not decoupled with word-level patterns.

\section{Related Work}

\textbf{Compositional generalization} (CG) has attracted much attention in NLP field.
Most existing benchmarks measured CG under fine-tuning with synthetic semantic parsing tasks, suggesting the limitations of general-purpose neural networks~\citep{lake2018generalization, keysers2019measuring, kim2020cogs}.
Many approaches were proposed to enhance the CG on general-purpose models~\citep{andreas2020good, akyurek2020learning, guo2021revisiting, oren2021finding, shaw2021compositional, zhu2021learning} or design task-specific methods~\citep{liu2020compositional, herzig2021span, chen2020compositional, liu2021learning}.
Some influential factors that affect CG have been revealed, such as the length bias~\citep{csordas2021devil}, target-side format~\citep{furrer2020compositional, herzig2021unlocking} and local structures~\citep{bogin2022unobserved}.
Most existing work explored CG under the fine-tuning paradigm, while our work advances the exploration under the in-context learning paradigm.

\vspace{2mm}
\noindent \textbf{In-context learning} (ICL) along with large language models (LLMs) has shown surprising performance in many NLP tasks~\citep{brown2020language, hendrycks2020measuring, patel2021mapping, rae2021scaling, zhang2022opt, hoffmann2022training, srivastava2022beyond, chowdhery2022palm, smith2022using, wei2022emergent}.
Most related to our work, \citet{qiu2022evaluating} and \citet{drozdov2022compositional} also explored ICL on CG challenges.
\citet{qiu2022evaluating} utilized the target-side similarity on structural fragments and reported that LLMs still exhibited much poorer CG than fine-tuned small models on COGS, which is close to our initial observations.
\citet{drozdov2022compositional} designed task-specific inference pipelines for performing CG under a least-to-most manner.
Our work provides more general understandings on how to improve CG performance by revealing several factors in selecting in-context examples.
In addition, some more recent work has similar observations on the potential of LLMs on CG~\citep{hosseini2022compositional}, gains from diversity~\citep{levy2022diverse}, and challenges under fictional words~\citep{kim2022uncontrolled}

\vspace{2mm}
\noindent \textbf{Selection of in-context examples} is an essential part for the utilization of ICL.
Most existing work considered the similarity as the major metric during selection.
\citet{liu2022makes} selected $k$-nearest neighbors with similar sentence embeddings;
\citet{shin2021constrained} regarded the conditional probability from a pre-trained LLM as the similarity score;
\citet{rubin2021learning} and \citet{zhang2022active} separately trained a retriever to score the similarity;
\citet{poesia2021synchromesh} and \citet{madaan2022language} estimated the target-side similarity.
This work demonstrates the necessity of structural similarity in achieving CG, and also reveals the importance of two other factors beyond similarity, i.e., diversity and complexity.

\section{Conclusion and Future Work}
This work investigates how in-context compositional generalization is affected by the selection of examples.
The test suite \textsc{CoFe} is constructed to study three factors.
Experiments show the effects of structural similarity, higher diversity and lower complexity.
Two challenges under in-context compositional generalization are further revealed.

To apply our revealed factors outside the \textsc{CoFe} test suite, one main challenge for future work is to determine the hidden structures behind expressions without knowing the exact generative grammar.
Here, we consider two potential approaches.
One is to use a pre-trained parser to generate a parse tree for the input query and then measure tree similarity.
The other approach is to pre-train an embedding model with a structure-aware training objective and then compute embedding similarity.

\section*{Limitations}

\paragraph{GPU resources.}
This work utilizes extremely large language models and thus has a high cost on GPU resources.
Concretely, experiments are conducted on the 8 x NVIDIA A100 GPU station.
The maximum inference time on each version of \textsc{CoFe} (containing 4,785 test cases) is $\sim8$ hours.
The maximum estimation of costed computing resources in this study is $\sim500$ x 8 GPU hours.

\paragraph{Synthetic data.}
As in most previous work on compositional generalization~\cite{lake2018generalization, keysers2019measuring, kim2020cogs}, the \textsc{CoFe} dataset is constructed using synthetic data rather than natural one.
The source-side sentences in \textsc{CoFe} are from COGS, which account for 70–80\% of naturally-occurring English sentences~\citep{kim2020cogs, roland2007frequency}.
Thus, this synthetic test suite could be close to the real-world application scenarios.

\paragraph{Single run.}
Due to the high cost on computing resources, we do not take multiple runs with different sets of examples, nor did we take multiple samples with temperature > 0.
Observations under different prompt orders (in Appendix~\ref{sec:robustness}) imply that with desired factors in selecting in-context examples, there could be low variance in experiments.

\section*{Ethics Statement}
Due to the utilization of pre-trained language models, this work could be exposed to some potential risks of ethical issues on general deep learning models (such as social bias and privacy breaches).
As explored in this work that the model behavior can be hugely influenced by the provided context, we call for further investigation into how ethical issues can be avoided by controlling the provided context.

\section*{Acknowledgments}
We thank all the anonymous reviewers for their valuable comments.
Shengnan An and Nanning Zheng were supported in part by NSFC under grant No. 62088102.

\bibliography{anthology,custom}
\bibliographystyle{acl_natbib}

\appendix

\clearpage

This is the Appendix of the paper: \textit{
How Do In-Context Examples Affect Compositional Generalization?}

\section{Grammar}\label{sec:ap_grammar}

\begin{table*}[h]
\caption{Part of the grammar used in constructing \textsc{CoFe}.
}
\label{tab:ap_grammar}
\centering
\resizebox{.99\linewidth}{!}{
\begin{tabular}{ll|l}
\toprule
Formal English Grammar & Semantic Representation & Type\\ \midrule
active-verb / passive-verb $\twoheadrightarrow$ $\mathbf{S}_{v}$ & PRED-FUNC $\twoheadrightarrow$ $\mathbf{S}_{P}$ & \multirow{5}{*}{T-Production Rule}\\
subject / direct-object / indirect-object $\twoheadrightarrow$ $\mathbf{S}_{n}$ & AGENT / THEME / RECIPIENT $\twoheadrightarrow$ $\mathbf{S}_{E}$ &\\
pp-mod / pp-s $\twoheadrightarrow$ $\mathbf{S}_{n}$ & PP-FUNC / PP-S $\twoheadrightarrow$ $\mathbf{S}_{E}$ &\\
conj $\twoheadrightarrow$ that & CP-CONCAT $\twoheadrightarrow$ CCOMP &\\
prep $\twoheadrightarrow$ in / on / beside & PP-CONCAT $\twoheadrightarrow$ IN / ON / BESIDE &\\
\midrule
sentence $\twoheadrightarrow$ subj active-verb 
& CLAUSE $\twoheadrightarrow$ PRED-FUNC ( AGENT, NONE, NONE ) & \multirow{3}{*}{N-Production Rule}\\
sentence $\twoheadrightarrow$ subj active-verb direct-obj indirect-obj
& CLAUSE $\twoheadrightarrow$ PRED-FUNC ( AGENT, THEME, RECIPIENT ) &\\
subject / direct-object / indirect-object $\twoheadrightarrow$ pp-mod & AGENT / THEME / RECIPIENT $\twoheadrightarrow$ PP-FUNC &\\
\midrule
sentence $\twoheadrightarrow$ sentence conj sentence & CLAUSE $\twoheadrightarrow$ CLAUSE CP-CONCAT CLAUSE & \multirow{2}{*}{R-Production Rule}\\
pp-mod $\twoheadrightarrow$ pp-s prep pp-mod & PP-FUNC $\twoheadrightarrow$ PP-CONCAT ( PP-S, PP-FUNC ) &\\
\bottomrule
\end{tabular}
}
\end{table*}

Part of the grammar used in constructing \textsc{CoFe} is listed in Table~\ref{tab:ap_grammar}.
Note that the max recursive times of R-Production Rules is 2 in prompting examples and 12 in test cases.
The target-side grammar follows the reconstruction in \citet{an2023does}.
Overall, the original target grammar of COGS is reconstructed to be chain-structured.
Concretely, first, the original output tokens in COGS are capitalized;
then, the variables (e.g., ``\textit{x\_1}'') in the original grammar are aligned and replaced with their corresponding terminals;
finally, the output clauses are grouped as the function format, in which the function name belongs to ``\textit{PRED-FUNC}'' and the arguments are ordered as ``\textit{AGENT}'', ``\textit{THEME}'', and ``\textit{RECIPIENT}''.
Moreover, if ``\textit{PRED-FUNC}'' does not contain one or some arguments, the positions of these arguments are filled with ``\textit{NONE}'' terminal.
For the two R-Production rules in Table~\ref{tab:ap_grammar}, the first is in chain structure and the second is in nested structure.
Moreover, the whole nested ``\textit{PP-FUNC}'' will be filled into the ``\textit{PRED-FUNC}'' as an argument, rather than concatenated to the tail of the ``\textit{CLAUSE}''.

\section{Details of Fine-Tuning}\label{sec:ap_models}

The fine-tuned GPT2-Large contains 762M parameters.
For fine-tuning, we take 50,000 training steps with 8 batch size and 1e-5 learning rate (without warm-up strategy).
We set weight decay as 1e-2 and label smoothing factor as 1e-1.
For inference with GPT2-Large, we set beam size as 5 and set max length as 1,024.

\section{Details of Implementation}\label{sec:ap_implement}

\subsection{Algorithm}\label{sec:ap_gsds}

\begin{algorithm}[t]
\caption{Greedy-Search Algorithm for Constructing \textsc{CoFe}}
\label{alg:ap_gsds}
\small
\renewcommand{\algorithmicrequire}{\textbf{Given:}}
\renewcommand{\algorithmicensure}{\textbf{Return:}}
\begin{algorithmic}[1]
\REQUIRE ~~\\
$(\mathbf{X}, \mathbf{Y})$: Source and target parse trees in one test case;\\
$\mathcal{B}$: Example bank;\\
$(\mathbf{X}_{i}, \mathbf{Y}_{i})\in\mathcal{B}$: One candidate case in example bank;\\
$\mathbf{X}_{A}$ and $\mathbf{Y}_{A}$: Aiming combination; \\
$w_p,w_s,w_c$: Weights for primitive coverage, structural similarity, and complexity penalty;\\
$\mathtt{P}(\cdot)$: primitives;\\
$\mathtt{S}(\cdot)$: structural combinations;\\

\ENSURE ~~\\
$C$: Selected in-context examples;

\STATE $C = \{\}$

\WHILE{$|C|<n$}
\STATE max\_score = 0
\STATE candidate = None

\FOR{$(\mathbf{X}_{i}, \mathbf{Y}_{i})\in \mathcal{B}$}
\STATE \textbf{Assert} $\mathbf{X}_{A}\notin\mathtt{S}(\mathbf{X}_{i})$
\STATE \textbf{Assert} $\mathbf{Y}_{A}\notin\mathtt{S}(\mathbf{Y}_{i})$
\STATE prim\_score = 0
\STATE stru\_score = 0

\FOR{element $\in\mathtt{P}(\mathbf{X}_{i})\cup\mathtt{P}(\mathbf{Y}_{i})$}
\IF{element $\in\mathtt{P}(\mathbf{X})\cup\mathtt{P}(\mathbf{Y})$}
\STATE prim\_score += $w_p$
\ENDIF
\ENDFOR

\FOR{element $\in\mathtt{S}(\mathbf{X}_{i})\cup\mathtt{S}(\mathbf{Y}_{i})$}
\IF{element $\in\mathtt{S}(\mathbf{X})\cup\mathtt{S}(\mathbf{Y})$ and element $\notin\mathtt{S}(C)$}
\STATE stru\_score += $w_s$
\ENDIF
\ENDFOR

comp\_penalty = $w_p \cdot \mathtt{depth}(\mathbf{X}_{i})$


score = prim\_score + stru\_score - comp\_penalty

\IF{score > max\_score}
\STATE max\_score = score
\STATE candidate = $(\mathbf{X}_{i}, \mathbf{Y}_{i})$
\ENDIF

\ENDFOR

\STATE $C$.add(candidate)

\ENDWHILE

\end{algorithmic}
\end{algorithm}

Algorithm~\ref{alg:ap_gsds} shows the greedy searching algorithm for constructing \textsc{CoFe}.

\subsection{Key Designs}

We give detailed descriptions of some key designs in Algorithm~\ref{alg:ap_gsds}.

\begin{itemize}
    \item $\mathtt{P}(\mathbf{T})$: Return the leaf nodes $\mathbf{T}_{L}$ on the tree;

    \item $\mathtt{S}(\mathbf{T})$: Return the structural combinations on the tree, i.e., $\mathbf{T}_{S}^{1}\cup\mathbf{T}_{S}^{>1}$;

    \item $w_p,w_s,w_c$: The initial scores for matching primitives, structural combinations, and complexity penalty, respectively.
    A higher $w$ means that the corresponding element is prioritized in greedy search.

    \item element $\notin \mathtt{S}(C)$: Already covered elements will not be awarded again, thus encouraging high diversity.

    \item $\mathtt{depth}(\mathbf{T})$: return the depth of the tree. Note that $\mathtt{depth}(\mathbf{X}_{i})=\mathtt{depth}(\mathbf{Y}_{i})$ in \textsc{CoFe}.
    
\end{itemize}

\subsection{Prompt Order}\label{sec:ap_order}
We take the structure-closer order, i.e., the examples in $C$ with a higher stru\_score are placed closer to the test case.
In Section~\ref{sec:robustness}, we show the robustness to the other two orders: random order, i.e., all selected in-context examples in $C$ are randomly shuffled, and atom-closer order, i.e., the examples in $C$ with a higher prim\_score are placed closer to the test case.

\subsection{Max Depth in $\mathbf{T}_{S}^{>1}$}
Since the max repetition times for \textit{LongChain} and \textit{DeepNest} are 2 (as described in Section~\ref{sec:cogs}), we set the max depth in $\mathbf{T}_{S}^{>1}$ as 2 in $\mathtt{S}(\mathbf{T})$.

\subsection{Similarity Under Diversity and Complexity Settings}\label{sec:statistics_diversity_complexity}

\begin{table}[h]
\caption{Statistics of different versions of \textsc{CoFe}  (\textit{PhraReco} category).
}
\label{tab:statistics_diversity_complexity}
\centering
\resizebox{.99\linewidth}{!}{
\begin{tabular}{@{}ccccc@{}}
\toprule
\multirow{2}{*}{Setting} & \multicolumn{4}{c}{Average Coverage} \\ \cmidrule(l){2-5} 
 & $\mathbf{T}_{L}$ & $\mathbf{T}_{N}$ & $\mathbf{T}_{S}^{1}$ & $\mathbf{T}_{S}^{>1}$ \\ \midrule
Default (Low Diversity, Mid   Complexity) & 100\% & 100\% & 84.4\% & 19.8\% \\
High Diversity & 100\% & 100\% & 84.4\% & 19.8\% \\
Low Complexity & 100\% & 100\% & 84.4\% & 19.8\% \\
High Complexity & 100\% & 100\% & 84.4\% & 19.8\% \\ \bottomrule
\end{tabular}
}
\end{table}

While changing diversity and complexity in variants of \textsc{CoFe} in Section~\ref{sec:diversity_complexity}, the primitive coverage and structural similarity are still satisfied.
Table~\ref{tab:statistics_diversity_complexity} shows that on\textit{PhraReco}, the statistics of coverage in different diversity and complexity settings are kept identical to the full similarity setting in \textsc{CoFe}.

\subsection{Excluding NL-Side Matching}\label{sec:ap_NL_similarity}

For excluding source-side matching in Section~\ref{sec:challenges}, besides removing the first term in Equation~\ref{equ:score}, we also limit the matching of $\mathbf{X}_{S}^{1}$.
Concretely, we require that the sentence rule in test case should not be covered by in-context examples.
The sentence rule is an N-Production rule that contains the non-terminal ``\textit{sentence}'' as the left hand.
To achieve this, we filter out test cases that can not meet this constraint.
Finally, 1,037 out of 4,785 test cases are kept in this variant of \textsc{CoFe}.

\subsection{Fictional Words}\label{sec:ap_familiarity}

For each target-side word that contain $l$ characters, we sequentially and randomly sample $l$ characters from alphabet as a fictional word to replace the original word.
In addition, for the experiments on fictional words, we take the atom-closer prompt order, since the model with this order performs better the default structure-closer order.

\section{Excluding Target-Side Matching}\label{sec:ap_only_source}

\begin{table*}[h]
\caption{Performances under only matching source side.
}
\label{tab:ap_only_source}
\centering
\resizebox{.8\linewidth}{!}{
\begin{tabular}{@{}cc|ccccc|c@{}}
\toprule
Model & Setting & PrimSubs & PrimAlte & PhraReco & LongChain & DeepNest & Average \\ \midrule
\multirow{2}{*}{code-davinci-002} & matching both side & 99.8 & 99.7 & 65.3 & 87.0 & 26.0 & 75.6 \\
 & only matching source side & 99.3 & 99.7 & 63.2 & 88.9 & 25.8 & 75.4 \\ \midrule
\multirow{2}{*}{text-davinci-002} & matching both side & 99.7 & 99.4 & 39.4 & 80.2 & 12.7 & 66.3 \\
 & only matching source side & 98.8 & 99.6 & 35.6 & 81.1 & 12.5 & 65.5 \\ \midrule
\multirow{2}{*}{code-cushman-002} & matching both side & 98.9 & 99.0 & 28.5 & 64.0 & 15.1 & 61.1 \\
 & only matching source side & 98.6 & 99.4 & 26.7 & 66.8 & 16.3 & 61.6 \\ \midrule
\multirow{2}{*}{code-cushman-001} & matching both side & 99.1 & 98.4 & 20.7 & 11.1 & 8.9 & 47.6 \\
 & only matching source side & 99.2 & 99.6 & 17.4 & 13.1 & 8.6 & 47.6 \\ \midrule
\multirow{2}{*}{davinci} & matching both side & 97.5 & 95.4 & 12.3 & 13.4 & 1.4 & 44.0 \\
 & only matching source side & 97.7 & 94.7 & 7.2 & 14.7 & 2.1 & 43.3 \\ \bottomrule
\end{tabular}
}
\end{table*}

In Section~\ref{sec:challenges}, we show that the performance drops with excluding the source-side matching.
Here, we examine the effect of target-side matching.
For constructing data, we directly remove the second term in Equation~\ref{equ:score}.
As shown in Table~\ref{tab:ap_only_source}, the performances with or without target-side matching are nearly identical.
Such an observation is similar to the comparison between oracle and non-oracle settings in \citet{qiu2022evaluating} that also utilized COGS benchmark, but different from \citet{poesia2021synchromesh} which suggested the importance of target-side similarity in code generation tasks.
We suppose there are mainly two reasons that could cause this difference.
On the one hand, different from general code generation tasks, the test suite for compositional generalization requires the exclusion of certain aiming combinations.
Therefore, the performance bottleneck in compositional generalization benchmarks mainly lies in the lacked aiming combinations.
On the other hand, in most compositional generalization benchmarks, the source-side matching could largely take over the target-side matching, since the terminals and rules in source grammar in these benchmarks are mapped many-to-one to the target grammar.
Therefore, when seeking for the source-side matching, the target-side matching is also improved.

\section{Illustration of Defined Notations}\label{sec:ap_notations}

\begin{figure*}[b]
    \centering
    \includegraphics[width=.95\textwidth]{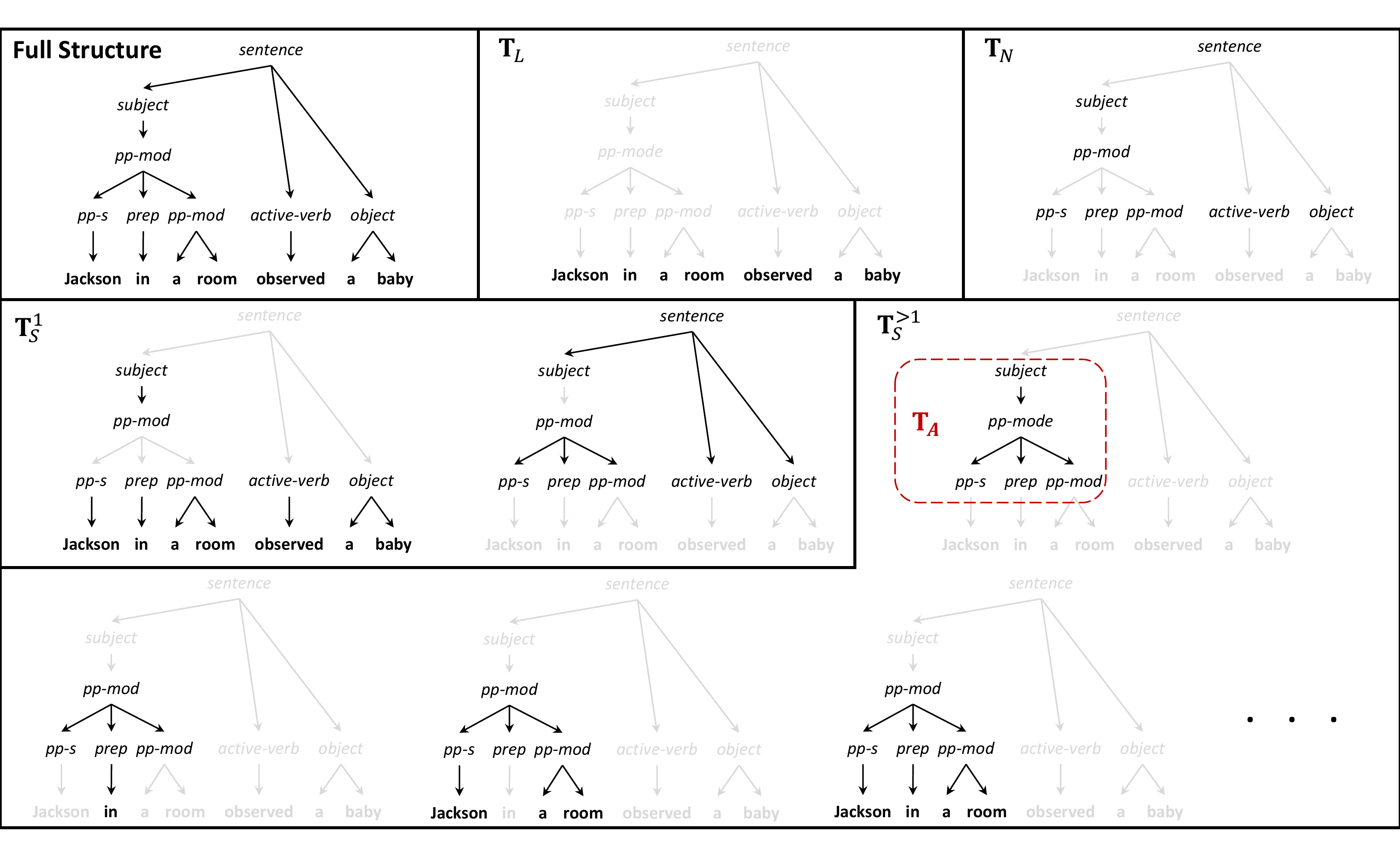}
    \caption{
    Illustration of defined notations.
    This test case belongs to \textit{PhraReco} category.
    $\mathbf{T}_{A}$ means the aiming combination in this test case.
    }
    \label{fig:ap_notations}
\end{figure*}

Figure~\ref{fig:ap_notations} illustrates the notations defined in Section~\ref{sec:formalize} based on a concrete expression ``\textit{Jackson in a room observed a baby}''.

Note that for all sub-structures in $\mathbf{T}_{S}^{1}\cup\mathbf{T}_{S}^{>1}$, we require them to be complete sub-structures.

\paragraph{Definition: Complete sub-structure (CSS).}
A CSS is a subgraph in a tree $\mathbf{T}$, satisfying that if an internal node in $\mathbf{T}$ and one of its child nodes are covered in this CSS, all other child nodes must be also covered in this CSS.

\section{Case Study}

We provide case study to further understanding the performance of compositional generalization observed in the main text.
For ease of reading, we include the following contents in the caption of figures.

\subsection{Two Types of Errors in \textit{DeepNest}}\label{sec:ap_case_study_deeper_nesting}

\begin{figure*}[t]
    \centering
    \includegraphics[width=.99\textwidth]{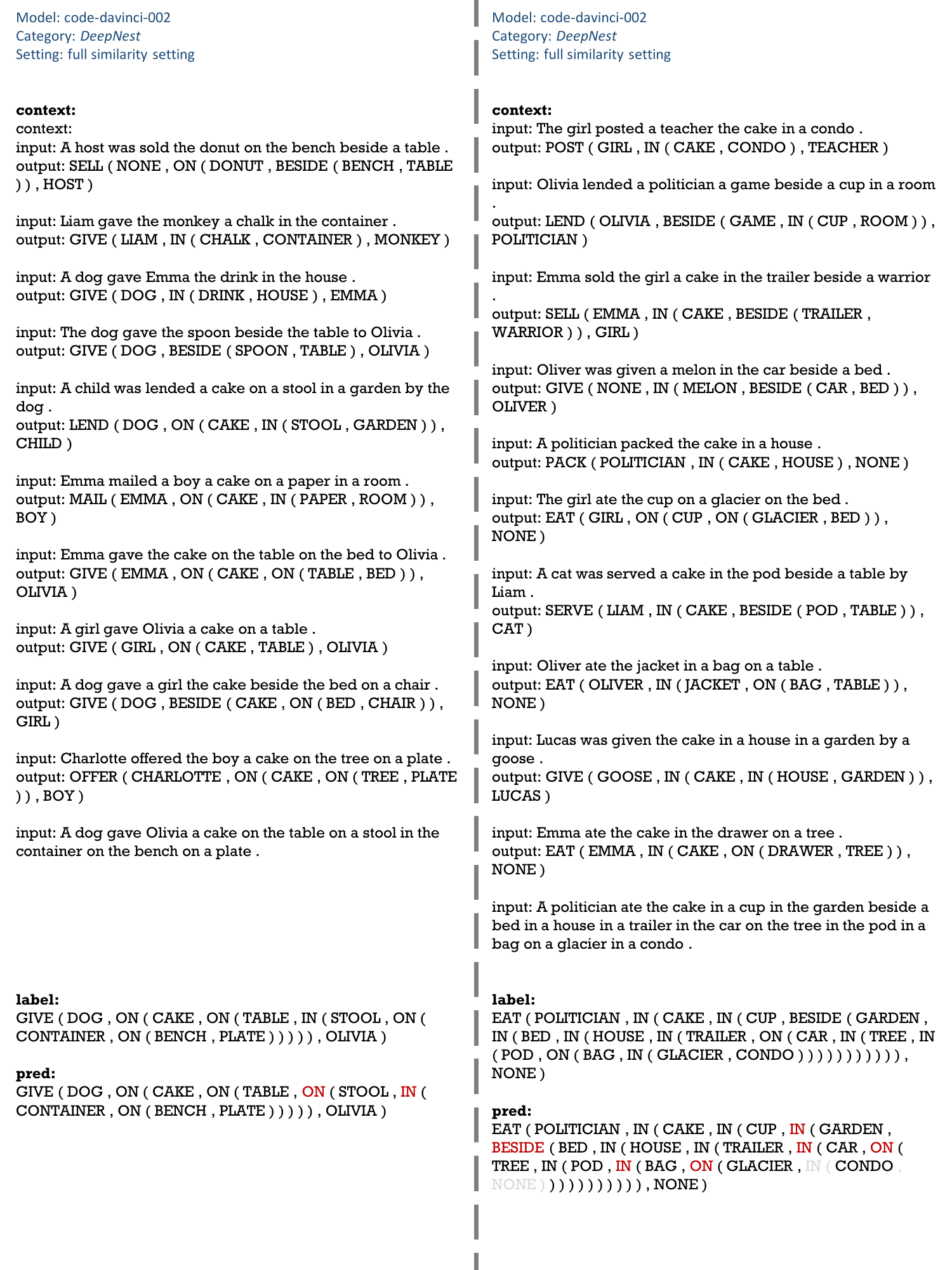}
    \caption{
    Two error cases in \textit{DeepNest} with code-davinci-002 model and full similarity setting.
    The overall structures of predictions in error cases are close to the ground truth, but the model makes mistakes on some local parts.
    Concretely, some local semantics are incorrect (in \textcolor[HTML]{C00000}{red}), and some words are redundant (in \textcolor[HTML]{D9D9D9}{gray}).
    }
    \label{fig:ap_case_study_deeper_nesting}
\end{figure*}

Figure~\ref{fig:ap_case_study_deeper_nesting} shows two error cases in \textit{DeepNest} with code-davinci-002 model and full similarity setting.
The overall structure of predictions are close to the ground truth, but the model makes mistakes on some local parts.
Concretely, some local semantics are incorrect (in \textcolor[HTML]{C00000}{red}), and some words are redundant (in \textcolor[HTML]{D9D9D9}{gray}).

Moreover, we also calculate the word-level coverage in predictions.
Besides the instance-level accuracy, we further investigate a word-level error rate on \textit{DeepNest}.
We find that in \textit{DeepNest}, 96.8\% of the words in the ground truth are contained by the predictions from code-davinci-002 (while only 48.8\% for GPT2-Large).
It indicates that the low instance-level accuracy is mainly caused by the wrong positions of words and redundant words.


\subsection{Structural Errors with Fictional Words}\label{sec:ap_case_study_novel_lexical_pattern}

\begin{figure*}[t]
    \centering
    \includegraphics[width=.99\textwidth]{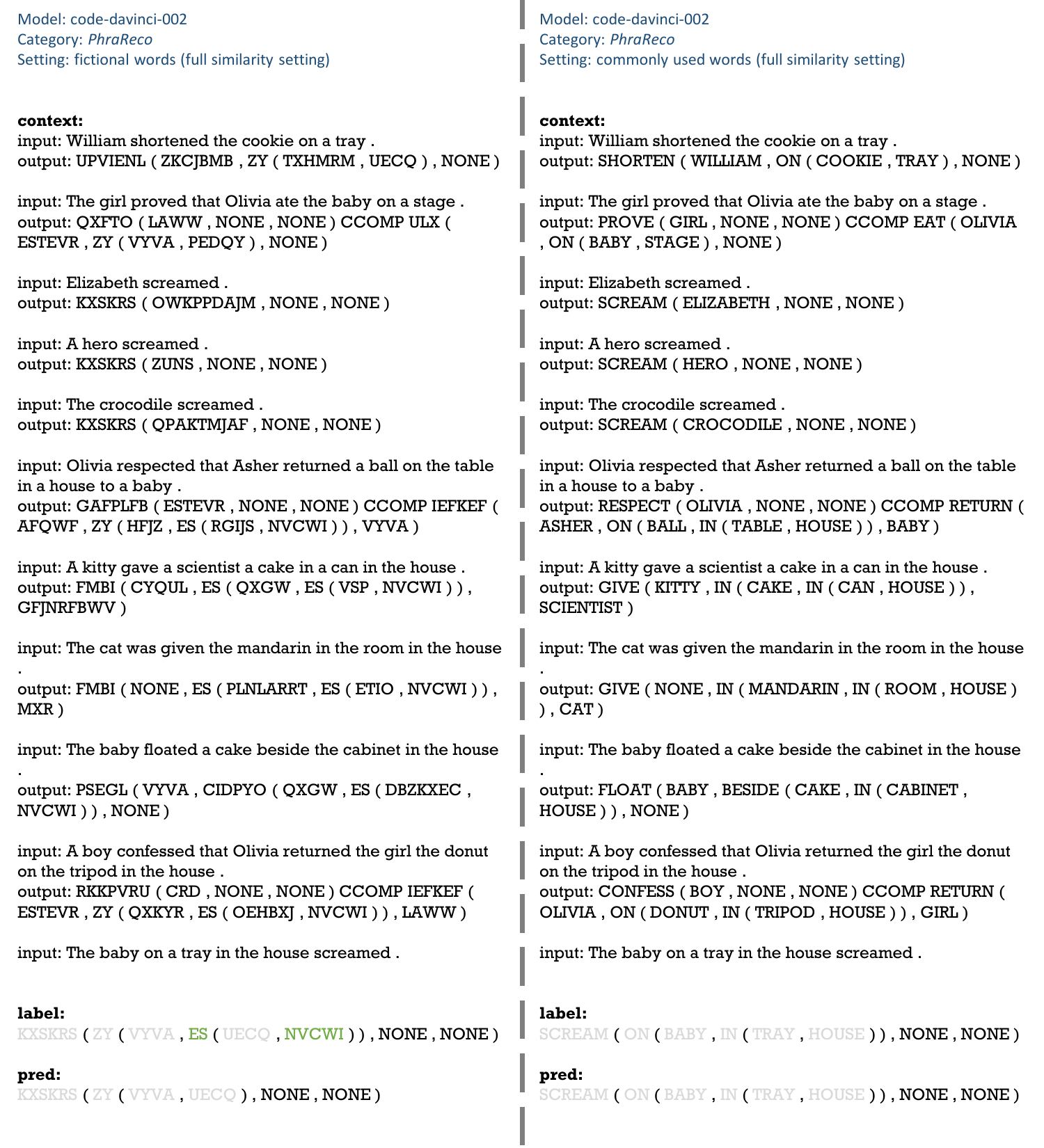}
    \caption{
    Comparison of performance between fictional words (left) and commonly used words (right).
    For the provided contexts on the left and right, the only difference is that the target-side words on the left are randomly selected characters while on the right they are uppercase of the source-side words.
    It shows that by changing only the target-side words, the model not only makes word-level errors (i.e., missing two words ``\textit{ES}'' and ``\textit{NVCWI}'' in prediction), it also generates the wrong parentheses structure (i.e., generate a 2-depth structure while in ground truth it is 3-depth).
    }
    \label{fig:ap_case_study_novel_lexical_pattern}
\end{figure*}

Figure~\ref{fig:ap_case_study_novel_lexical_pattern} shows the comparison of performance between fictional words (left) and commonly used words (right).
For the provided contexts on the left and right, the only difference is that the target-side words on the left are randomly selected characters while on the right they are uppercase of the source-side words.
It shows that by changing only the target-side words, the model not only makes word-level errors (i.e., missing two words ``\textit{ES}'' and ``\textit{NVCWI}'' in prediction), it also generates the wrong parentheses structure (i.e., generate a 2-depth structure while in ground truth it is 3-depth).

\subsection{Fail to Recognize Semantic Equivalence}\label{sec:ap_case_study_semantic_equivalence}

\begin{figure*}[t]
    \centering
    \includegraphics[width=.99\textwidth]{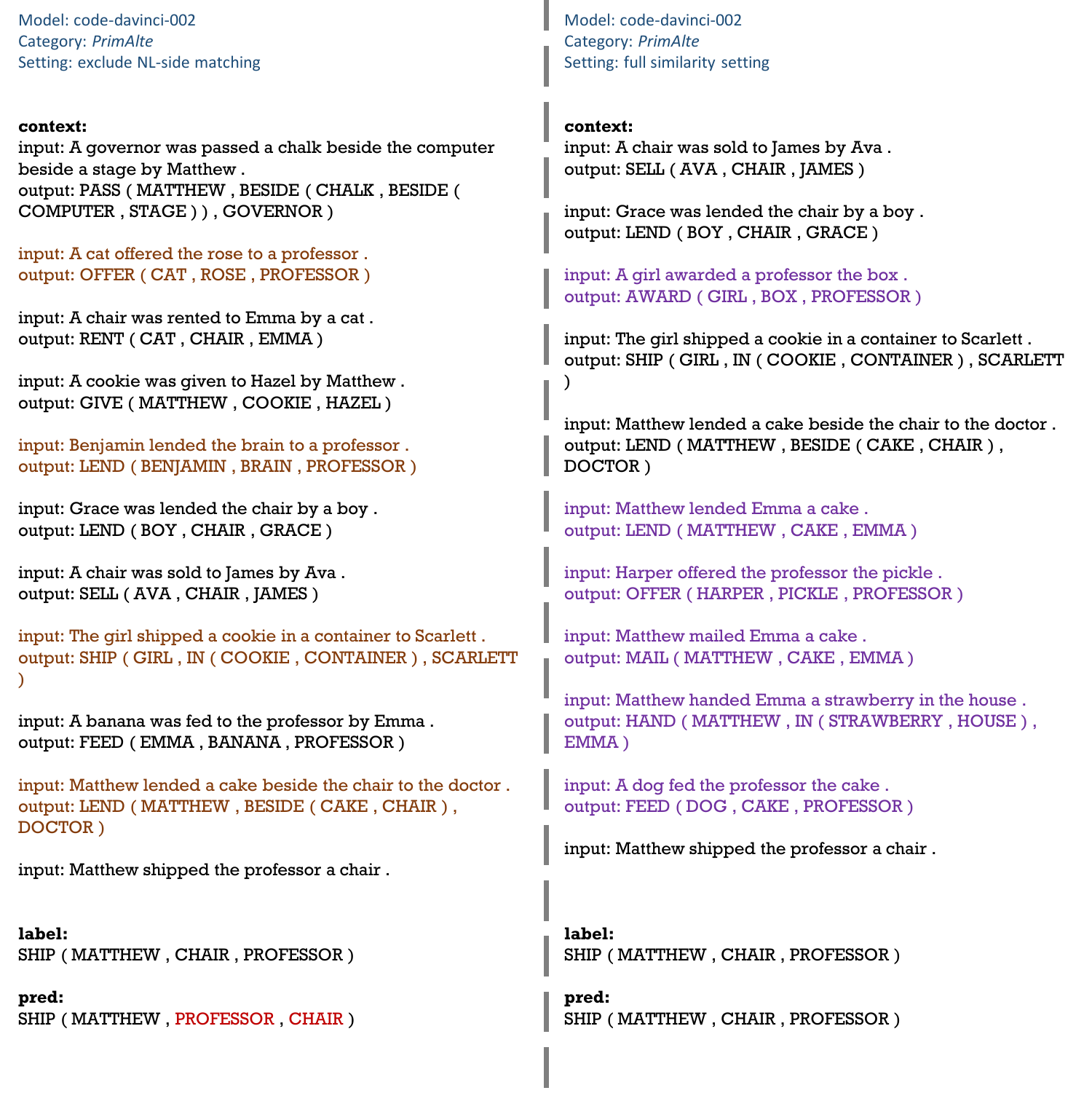}
    \caption{
    Comparison of performances between excluding NL-side matching (left) and containing NL-side matching (right).
    For the test input ``\textit{Matthew shipped the professor a chair .}'', it contains the sentence structure ``\textit{subject verb object\_1 object\_2}'' behind the NL expression.
    Context on the left does not explicitly contain this sentence structure,  but it contains a semantically equivalent structure (i.e., ``\textit{subject verb object\_2 to object\_1}'').
    However, the model generates the correct prediction on the right while fails on the left.
    Concretely, according to the wrong prediction on the left, the model perhaps considers that the semantics of ``\textit{subject verb object\_1 object\_2}'' is equivalent with ``\textit{subject verb object\_1 to object\_2}''.
    }
    \label{fig:ap_case_study_semantic_equivalence}
\end{figure*}

Figure~\ref{fig:ap_case_study_semantic_equivalence} shows the comparison of performances between excluding NL-side matching (left) and containing NL-side matching (right).
For the test input ``\textit{Matthew shipped the professor a chair .}'', it contains the sentence structure ``\textit{subject verb object\_1 object\_2}'' behind the NL expression.
Context on the left does not explicitly contain this sentence structure,  but it contains a semantically equivalent structure (i.e., ``\textit{subject verb object\_2 to object\_1}'').
However, the model generates the correct prediction on the right while fails on the left.
Concretely, according to the wrong prediction on the left, the model perhaps considers that the semantics of ``\textit{subject verb object\_1 object\_2}'' is equivalent with ``\textit{subject verb object\_1 to object\_2}''.

\subsection{Low Diversity Block Generalization}\label{sec:ap_case_study_diversity}

\begin{figure*}[t]
    \centering
    \includegraphics[width=.99\textwidth]{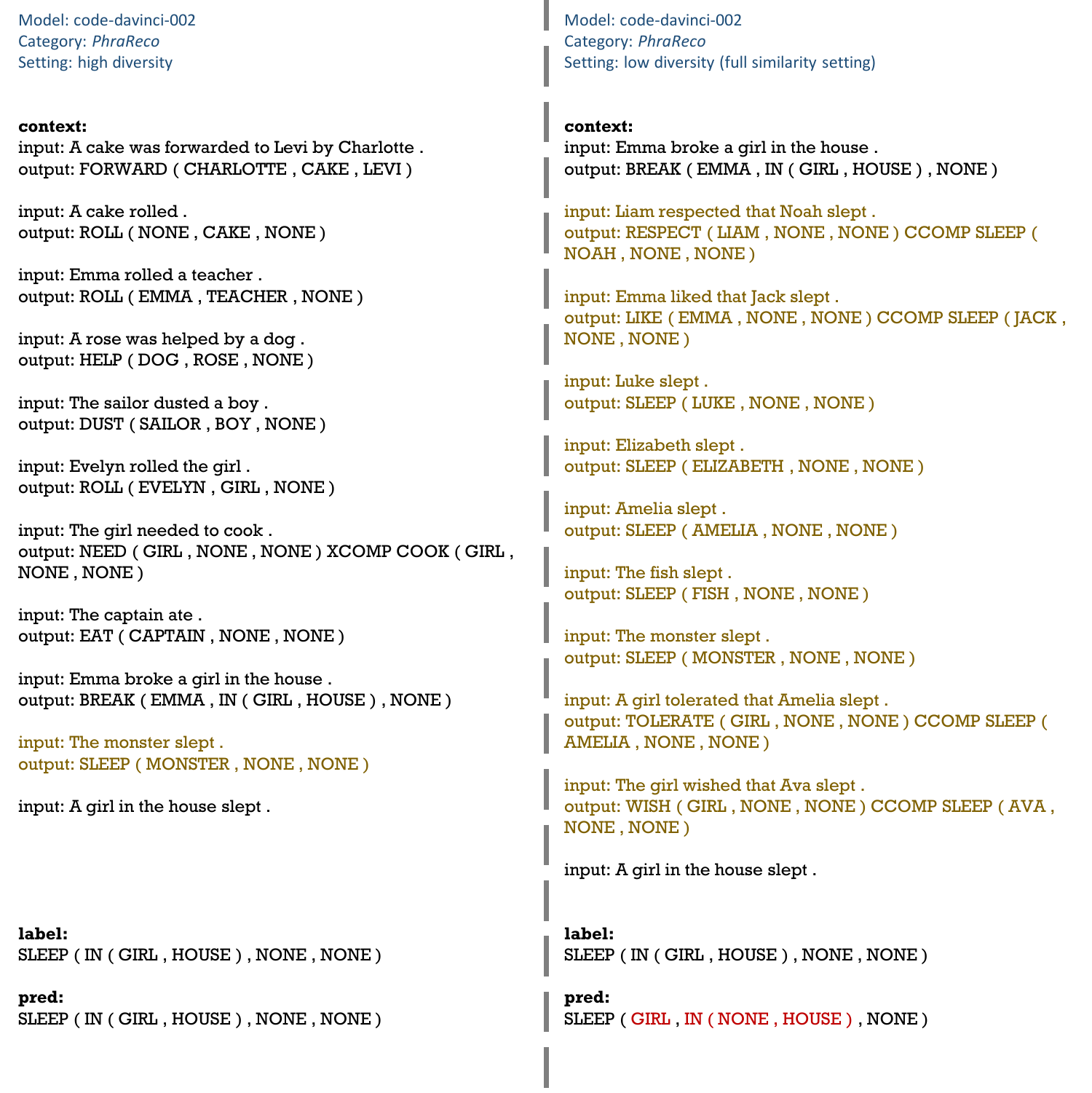}
    \caption{
    Comparison of performances on \textit{PhraReco} under high diversity (left) and low diversity (right).
    For the test input ``\textit{A girl in the house slept}'', ``\textit{subject slept}'' is one element contained in $\mathbf{T}_{S}^{>1}$.
    This element is repeatedly covered in the context on the right (low diversity) while only covered once on the left (high diversity).
    However, under high repetitiveness, the model fails on the test case, but succeed when there is low repetitiveness.
    }
    \label{fig:ap_case_study_diversity}
\end{figure*}

Figure~\ref{fig:ap_case_study_diversity} shows the comparison of performances on \textit{PhraReco} under high diversity (left) and low diversity (right).
For the test input ``\textit{A girl in the house slept}'', ``\textit{subject slept}'' is one element contained in $\mathbf{T}_{S}^{>1}$.
This element is repeatedly covered in the context on the right (low diversity) while only covered once on the left (high diversity).
However, under high repetitiveness, the model fails on the test case, but succeed when there is low repetitiveness.

\subsection{High Complexity Block Generalization}\label{sec:ap_case_study_complexity}

\begin{figure*}[t]
    \centering
    \includegraphics[width=.99\textwidth]{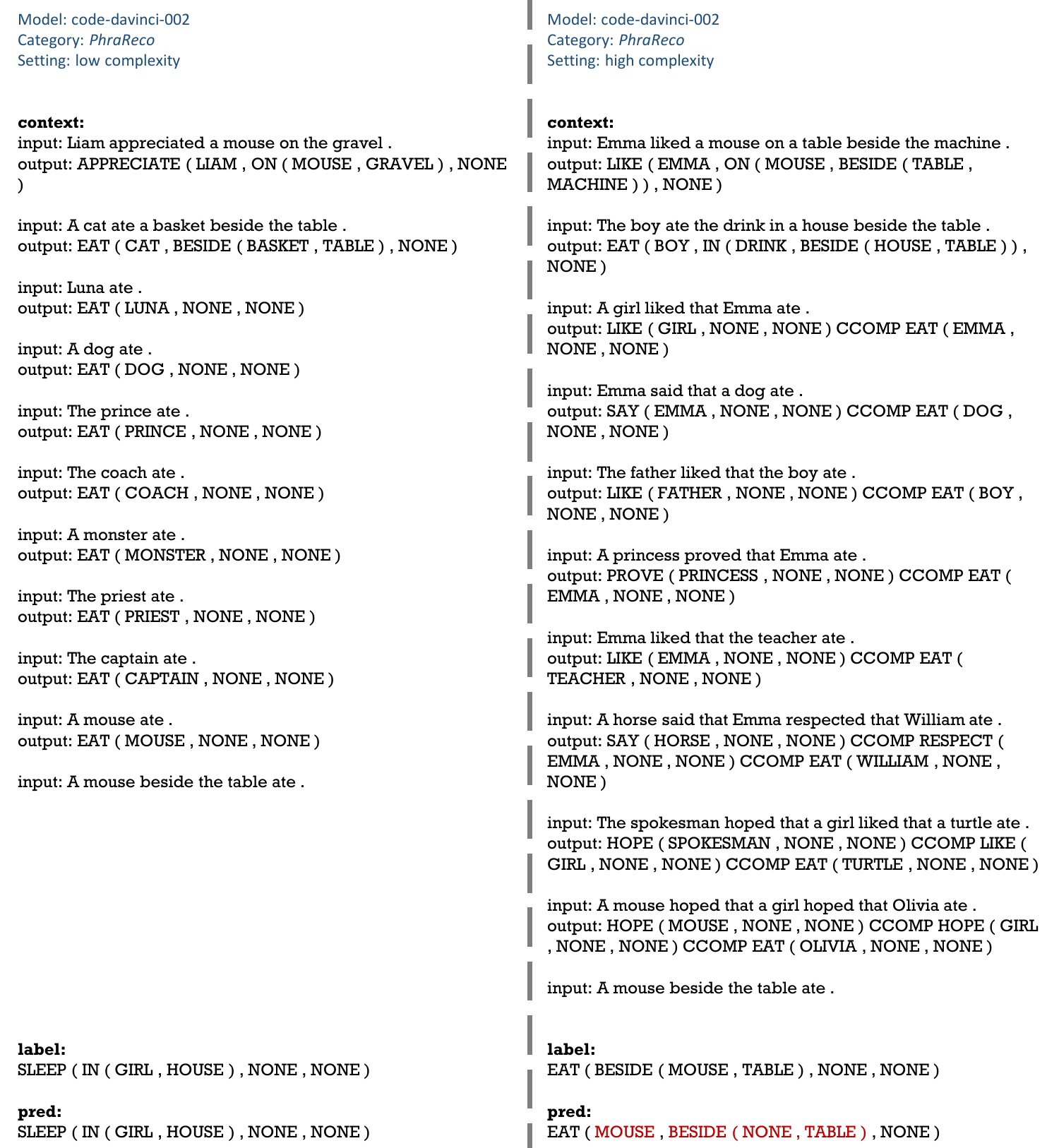}
    \caption{
    Comparison of performance on \textit{PhraReco} under low complexity (left) and high diversity (right).
    With low complexity, the test case is covered by simple and short in-context examples, and the model succeeds on the test case.
    With high complexity, the test case is covered by more complex and longer examples, and the model fails on the test case.
    }
    \label{fig:ap_case_study_complexity}
\end{figure*}

Figure~\ref{fig:ap_case_study_complexity} shows the comparison of performance on \textit{PhraReco} under low complexity (left) and high diversity (right).
With low complexity, the test case is covered by simple and short in-context examples, and the model succeeds on the test case.
With high complexity, the test case is covered by more complex and longer examples, and the model fails on the test case.

\section{Full Results}\label{sec:ap_full_results}

Due to the page limitation for main text, here we list our full results in Section~\ref{sec:experiments_and_analysis}.
The results in \textit{Assembling} are the best performance under each category among all combinations of factors.

\begin{table*}[b]
\caption{Full results.
}
\label{tab:ap_complete}
\centering
\resizebox{.99\linewidth}{!}{
\begin{tabular}{@{}c|cccccccc|ccccc|c@{}}
\toprule
\multirow{2}{*}{Model} & \multicolumn{1}{c|}{\multirow{2}{*}{Primitive}} & \multicolumn{2}{c|}{Similarity} & \multicolumn{2}{c|}{Diversity} & \multicolumn{3}{c|}{Complexity} & \multirow{2}{*}{PrimSubs} & \multirow{2}{*}{PrimAlte} & \multirow{2}{*}{PhraReco} & \multirow{2}{*}{LongChain} & \multirow{2}{*}{DeepNest} & \multirow{2}{*}{Average} \\
 & \multicolumn{1}{c|}{} & Rough & \multicolumn{1}{c|}{Precise} & Low & \multicolumn{1}{c|}{High} & Low & Mid & High &  &  &  &  &  &  \\ \midrule
\multirow{7}{*}{code-davinci-002} & \multicolumn{1}{c|}{\checkmark} &  & \multicolumn{1}{c|}{} &  & \multicolumn{1}{c|}{} &  &  &  & 92.2 & 77.1 & 60.8 & 62.1 & 12.3 & 60.9 \\ \cmidrule(l){2-15} 
 & \multicolumn{1}{c|}{\checkmark} & \checkmark & \multicolumn{1}{c|}{\checkmark} & \checkmark & \multicolumn{1}{c|}{} &  & \checkmark &  & 99.8 & 99.7 & 65.3 & 87.0 & 26.0 & 75.6 \\ \cmidrule(l){2-15} 
 & \multicolumn{1}{c|}{\checkmark} & \checkmark & \multicolumn{1}{c|}{} & \checkmark & \multicolumn{1}{c|}{} &  & \checkmark &  & 97.7 & 92.1 & 77.6 & 80.4 & 18.3 & 73.2 \\ \cmidrule(l){2-15} 
 & \multicolumn{1}{c|}{\checkmark} & \checkmark & \multicolumn{1}{c|}{\checkmark} &  & \multicolumn{1}{c|}{\checkmark} &  & \checkmark &  & - & - & 80.0 & 87.6 & 26.2 & 64.6 \\ \cmidrule(l){2-15} 
 & \multicolumn{1}{c|}{\checkmark} & \checkmark & \multicolumn{1}{c|}{\checkmark} & \checkmark & \multicolumn{1}{c|}{} & \checkmark &  &  & - & - & 67.6 & 87.3 & 25.6 & 60.2 \\ \cmidrule(l){2-15} 
 & \multicolumn{1}{c|}{\checkmark} & \checkmark & \multicolumn{1}{c|}{\checkmark} & \checkmark & \multicolumn{1}{c|}{} &  &  & \checkmark & - & - & 56.9 & 87.6 & 26.0 & 56.8 \\ \cmidrule(l){2-15} 
 & \multicolumn{8}{c|}{Assembling Desired   Factors} & 99.8 & 99.7 & 80.0 & 87.6 & 26.2 & 78.7 \\ \midrule
\multirow{7}{*}{text-chat-davinci-002} & \multicolumn{1}{c|}{\checkmark} &  & \multicolumn{1}{c|}{} &  & \multicolumn{1}{c|}{} &  &  &  & 92.2 & 75.4 & 47.0 & 65.0 & 6.3 & 57.2 \\ \cmidrule(l){2-15} 
 & \multicolumn{1}{c|}{\checkmark} & \checkmark & \multicolumn{1}{c|}{\checkmark} & \checkmark & \multicolumn{1}{c|}{} &  & \checkmark &  & 99.5 & 99.3 & 53.4 & 87.7 & 18.9 & 71.8 \\ \cmidrule(l){2-15} 
 & \multicolumn{1}{c|}{\checkmark} & \checkmark & \multicolumn{1}{c|}{} & \checkmark & \multicolumn{1}{c|}{} &  & \checkmark &  & 96.1 & 89.7 & 62.9 & 80.1 & 11.7 & 68.1 \\ \cmidrule(l){2-15} 
 & \multicolumn{1}{c|}{\checkmark} & \checkmark & \multicolumn{1}{c|}{\checkmark} &  & \multicolumn{1}{c|}{\checkmark} &  & \checkmark &  & - & - & 69.2 & 87.6 & 18.2 & 58.3 \\ \cmidrule(l){2-15} 
 & \multicolumn{1}{c|}{\checkmark} & \checkmark & \multicolumn{1}{c|}{\checkmark} & \checkmark & \multicolumn{1}{c|}{} & \checkmark &  &  & - & - & 55.1 & 87.6 & 19.0 & 53.9 \\ \cmidrule(l){2-15} 
 & \multicolumn{1}{c|}{\checkmark} & \checkmark & \multicolumn{1}{c|}{\checkmark} & \checkmark & \multicolumn{1}{c|}{} &  &  & \checkmark & - & - & 45.1 & 88.2 & 19.2 & 50.8 \\ \cmidrule(l){2-15} 
 & \multicolumn{8}{c|}{Assembling Desired   Factors} & 99.5 & 99.3 & 69.2 & 88.2 & 19.2 & 75.1 \\ \midrule
\multirow{7}{*}{text-davinci-002} & \multicolumn{1}{c|}{\checkmark} &  & \multicolumn{1}{c|}{} &  & \multicolumn{1}{c|}{} &  &  &  & 88.5 & 66.4 & 38.7 & 46.5 & 2.9 & 48.6 \\ \cmidrule(l){2-15} 
 & \multicolumn{1}{c|}{\checkmark} & \checkmark & \multicolumn{1}{c|}{\checkmark} & \checkmark & \multicolumn{1}{c|}{} &  & \checkmark &  & 99.7 & 99.4 & 39.4 & 80.2 & 12.7 & 66.3 \\ \cmidrule(l){2-15} 
 & \multicolumn{1}{c|}{\checkmark} & \checkmark & \multicolumn{1}{c|}{} & \checkmark & \multicolumn{1}{c|}{} &  & \checkmark &  & 94.9 & 86.7 & 55.9 & 66.3 & 8.1 & 62.4 \\ \cmidrule(l){2-15} 
 & \multicolumn{1}{c|}{\checkmark} & \checkmark & \multicolumn{1}{c|}{\checkmark} &  & \multicolumn{1}{c|}{\checkmark} &  & \checkmark &  & - & - & 60.6 & 78.7 & 12.3 & 50.5 \\ \cmidrule(l){2-15} 
 & \multicolumn{1}{c|}{\checkmark} & \checkmark & \multicolumn{1}{c|}{\checkmark} & \checkmark & \multicolumn{1}{c|}{} & \checkmark &  &  & - & - & 43.2 & 79.9 & 12.9 & 45.3 \\ \cmidrule(l){2-15} 
 & \multicolumn{1}{c|}{\checkmark} & \checkmark & \multicolumn{1}{c|}{\checkmark} & \checkmark & \multicolumn{1}{c|}{} &  &  & \checkmark & - & - & 33.5 & 80.2 & 12.8 & 42.2 \\ \cmidrule(l){2-15} 
 & \multicolumn{8}{c|}{Assembling Desired   Factors} & 99.7 & 99.4 & 60.6 & 80.2 & 12.9 & 70.6 \\ \midrule
\multirow{7}{*}{code-cushman-002} & \multicolumn{1}{c|}{\checkmark} &  & \multicolumn{1}{c|}{} &  & \multicolumn{1}{c|}{} &  &  &  & 82.6 & 55.6 & 21.3 & 29.3 & 5.0 & 38.8 \\ \cmidrule(l){2-15} 
 & \multicolumn{1}{c|}{\checkmark} & \checkmark & \multicolumn{1}{c|}{\checkmark} & \checkmark & \multicolumn{1}{c|}{} &  & \checkmark &  & 98.9 & 99.0 & 28.5 & 64.0 & 15.1 & 61.1 \\ \cmidrule(l){2-15} 
 & \multicolumn{1}{c|}{\checkmark} & \checkmark & \multicolumn{1}{c|}{} & \checkmark & \multicolumn{1}{c|}{} &  & \checkmark &  & 94.0 & 77.7 & 31.4 & 44.7 & 10.3 & 51.6 \\ \cmidrule(l){2-15} 
 & \multicolumn{1}{c|}{\checkmark} & \checkmark & \multicolumn{1}{c|}{\checkmark} &  & \multicolumn{1}{c|}{\checkmark} &  & \checkmark &  & - & - & 40.8 & 62.4 & 14.9 & 39.4 \\ \cmidrule(l){2-15} 
 & \multicolumn{1}{c|}{\checkmark} & \checkmark & \multicolumn{1}{c|}{\checkmark} & \checkmark & \multicolumn{1}{c|}{} & \checkmark &  &  & - & - & 31.9 & 64.3 & 15.8 & 37.3 \\ \cmidrule(l){2-15} 
 & \multicolumn{1}{c|}{\checkmark} & \checkmark & \multicolumn{1}{c|}{\checkmark} & \checkmark & \multicolumn{1}{c|}{} &  &  & \checkmark & - & - & 22.6 & 64.5 & 14.6 & 33.9 \\ \cmidrule(l){2-15} 
 & \multicolumn{8}{c|}{Assembling Desired   Factors} & 98.9 & 99.0 & 40.8 & 64.5 & 15.8 & 63.8 \\ \midrule
\multirow{7}{*}{code-cushman-001} & \multicolumn{1}{c|}{\checkmark} &  & \multicolumn{1}{c|}{} &  & \multicolumn{1}{c|}{} &  &  &  & 76.6 & 60.7 & 16.9 & 5.0 & 1.0 & 32.0 \\ \cmidrule(l){2-15} 
 & \multicolumn{1}{c|}{\checkmark} & \checkmark & \multicolumn{1}{c|}{\checkmark} & \checkmark & \multicolumn{1}{c|}{} &  & \checkmark &  & 99.1 & 98.4 & 20.7 & 11.1 & 8.9 & 47.6 \\ \cmidrule(l){2-15} 
 & \multicolumn{1}{c|}{\checkmark} & \checkmark & \multicolumn{1}{c|}{} & \checkmark & \multicolumn{1}{c|}{} &  & \checkmark &  & 92.5 & 86.0 & 24.7 & 8.0 & 3.5 & 42.9 \\ \cmidrule(l){2-15} 
 & \multicolumn{1}{c|}{\checkmark} & \checkmark & \multicolumn{1}{c|}{\checkmark} &  & \multicolumn{1}{c|}{\checkmark} &  & \checkmark &  & - & - & 31.4 & 12.8 & 8.4 & 17.5 \\ \cmidrule(l){2-15} 
 & \multicolumn{1}{c|}{\checkmark} & \checkmark & \multicolumn{1}{c|}{\checkmark} & \checkmark & \multicolumn{1}{c|}{} & \checkmark &  &  & - & - & 23.2 & 12.7 & 8.9 & 14.9 \\ \cmidrule(l){2-15} 
 & \multicolumn{1}{c|}{\checkmark} & \checkmark & \multicolumn{1}{c|}{\checkmark} & \checkmark & \multicolumn{1}{c|}{} &  &  & \checkmark & - & - & 18.6 & 11.5 & 8.7 & 12.9 \\ \cmidrule(l){2-15} 
 & \multicolumn{8}{c|}{Assembling Desired   Factors} & 99.1 & 98.4 & 31.4 & 12.8 & 8.9 & 50.1 \\ \midrule
\multirow{7}{*}{code-cushman-001} & \multicolumn{1}{c|}{\checkmark} &  & \multicolumn{1}{c|}{} &  & \multicolumn{1}{c|}{} &  &  &  & 69.4 & 52.3 & 9.4 & 2.3 & 0.2 & 26.7 \\ \cmidrule(l){2-15} 
 & \multicolumn{1}{c|}{\checkmark} & \checkmark & \multicolumn{1}{c|}{\checkmark} & \checkmark & \multicolumn{1}{c|}{} &  & \checkmark &  & 97.5 & 95.4 & 12.3 & 13.4 & 1.4 & 44.0 \\ \cmidrule(l){2-15} 
 & \multicolumn{1}{c|}{\checkmark} & \checkmark & \multicolumn{1}{c|}{} & \checkmark & \multicolumn{1}{c|}{} &  & \checkmark &  & 79.4 & 66.6 & 18.8 & 4.3 & 1.3 & 34.1 \\ \cmidrule(l){2-15} 
 & \multicolumn{1}{c|}{\checkmark} & \checkmark & \multicolumn{1}{c|}{\checkmark} &  & \multicolumn{1}{c|}{\checkmark} &  & \checkmark &  & - & - & 20.0 & 10.2 & 1.3 & 10.5 \\ \cmidrule(l){2-15} 
 & \multicolumn{1}{c|}{\checkmark} & \checkmark & \multicolumn{1}{c|}{\checkmark} & \checkmark & \multicolumn{1}{c|}{} & \checkmark &  &  & - & - & 14.7 & 13.8 & 1.4 & 10.0 \\ \cmidrule(l){2-15} 
 & \multicolumn{1}{c|}{\checkmark} & \checkmark & \multicolumn{1}{c|}{\checkmark} & \checkmark & \multicolumn{1}{c|}{} &  &  & \checkmark & - & - & 7.8 & 13.5 & 1.3 & 7.5 \\ \cmidrule(l){2-15} 
 & \multicolumn{8}{c|}{Assembling Desired   Factors} & 97.5 & 95.4 & 20.0 & 13.8 & 1.4 & 45.6 \\ \midrule
\multicolumn{1}{l|}{Fine-Tuned GPT2-Large} & \multicolumn{8}{c|}{-} & 93.6 & 97.9 & 14.0 & 5.4 & 0.0 & 42.2 \\ \bottomrule
\end{tabular}
}
\end{table*}

\end{document}